\documentclass[10pt,journal,compsoc]{IEEEtran}

\usepackage{graphicx, epstopdf, amsmath, float, amssymb, amsthm, mathtools}
\usepackage{subfig}
\usepackage[ruled]{algorithm2e}
\def\R{\mathbb{R}}
\def\AB{\allowbreak}

\usepackage{url}
\usepackage[usenames,dvipsnames]{color}
\usepackage[colorlinks=true]{hyperref}
\hypersetup{
  colorlinks,
 citecolor=Blue,
 linkcolor=Red,
 urlcolor=Violet}

\ifCLASSOPTIONcompsoc
  \usepackage[nocompress]{cite}
\else
  \usepackage{cite}
\fi
\usepackage{array}
\begin{document}
\title{Joint Color-Spatial-Directional clustering and Region Merging (JCSD-RM) for unsupervised RGB-D image segmentation}
\author{Md.~Abul~Hasnat,
        Olivier~Alata
        and~Alain~Tr{\'e}meau
\thanks{Laboratoire Hubert Curien, Universit{\'e} Jean Monnet, Saint Etienne, France.}%
\thanks{e-mail:$\lbrace$mdabul.hasnat, olivier.alata, alain.tremeau$\rbrace$@univ-st-etienne.fr.}}
\IEEEtitleabstractindextext{%
\begin{abstract}
Recent advances in depth imaging sensors provide easy access to the synchronized depth with color, called RGB-D image. In this paper, we propose an unsupervised method for indoor RGB-D image segmentation and analysis. We consider a statistical image generation model based on the color and geometry of the scene.  Our method consists of a joint color-spatial-directional clustering method followed by a statistical planar region merging method. We evaluate our method on the NYU depth database and compare it with existing unsupervised RGB-D segmentation methods. Results show that, it is comparable with the state of the art methods and it needs less computation time. Moreover, it opens interesting perspectives to fuse color and geometry in an unsupervised manner.
\end{abstract}

}

\maketitle
\IEEEdisplaynontitleabstractindextext
\IEEEpeerreviewmaketitle
\section{Introduction}
\label{sec:introduction}

\IEEEPARstart{I}{n} the field of image processing, segmentation is considered as one of the oldest and most widely studied problems that groups perceptually similar pixels based on certain features, e.g., color, texture, etc. A variety of different techniques already exist in literature \cite{szeliski2011computer}, which address image segmentation from different perspectives. In this paper, we address the problem of synchronized color and depth image segmentation and propose a solution that combines a clustering method \cite{murphy2012machine} with a statistical region merging technique \cite{nock2004statistical}.

After the introduction of Microsoft Kinect camera in 2010, the availability of RGB-D images is widespread now \cite{zhang2012microsoft, han2013enhanced}. As a consequence, traditional computer vision algorithms which had been previously developed for color/intensity image, have been enhanced to incorporate depth information \cite{han2013enhanced}. Recent progresses on RGB-D image segmentation  \cite{gupta2013perceptual, taylor2013parsing, silberman2012indoor, ren2012rgb, dal2012fusion, koppula2011semantic} have shown that depth as an additional feature improves accuracy of this task. Most of the techniques address the problem with supervised approaches (e.g., \cite{gupta2013perceptual}). In contrary, unsupervised approach (e.g., \cite{dal2012fusion}) remains underexplored. Moreover, it remains an important issue - what is the best way to fuse color and geometry in an unsupervised manner? This motivates us to conduct further research and contribute towards unsupervised indoor RGB-D image segmentation or scene labeling.

This paper proposes a scene segmentation approach which first identifies the possible image regions w.r.t. a statistical image generation model. Then it merges regions using the planar statistics and the RGB-D image gradients. The proposed model is based on three different cues/features of the RGB-D image: color, 3D location and surface normals. It follows a generative model-based approach for these features in which they are issued independently (\emph{“na\"{\i}ve” Bayes} \cite{zhang2004optimality, lewis1998naive} assumption) from a finite mixture of certain probability distributions. The model considers the Gaussian distribution \cite{murphy2012machine} for color and 3D features and the directional (Fisher or Watson) distribution \cite{mardia2009directional, hasnat2015mbhcfmm, hasnat2014wmm} for surface normals. We use the directional distribution because: (a) it provides adequate statistics to explain the planar property of regions and (b) it helps us to develop a simple and effective region merging method. A common property of the Gaussian and directional (Fisher or Watson) distributions is that they belong to the Regular Exponential Family (REF) \cite{liu2012shape, nielsen2009statistical, mardia2009directional}. We exploit this property to develop an efficient clustering method based on the proposed image generation model.

Finite Mixture Models are commonly used for cluster analysis \cite{fraley1998many, biernacki2000assessing, fraley2007model, figueiredo2002unsupervised}. In the context of image analysis and segmentation these models have been employed with the Gaussian distribution for clustering the color image pixels \cite{ma2007segmentation, alata2009there, garcia2010simplification, szeliski2011computer, nguyen2013fast}. Recently, these models have been exploited to analyze 3D images by clustering the surface normals with a mixture of directional distributions \cite{hasnat2015mbhcfmm, hasnat2014wmm, hasnat2013hierarchical}. These clusters are obtained by using the Expectation Maximization (EM) algorithm that performs Maximum Likelihood Estimate (MLE) of the model parameters \cite{mclachlan2008em, murphy2012machine, bishop2006pattern}. This paper proposes an EM method for a combined mixture model of multiple probability distributions, where each distribution belongs to the REF. Precisely, we propose an EM method for joint clustering of independent features.

Bregman Soft Clustering (BSC) is a centroid based parametric clustering method \cite{banerjee2005clusteringb}. It has been effectively employed to estimate parameters of mixture models which are based on the REF \cite{nielsen2009statistical}. Compare to the traditional EM based algorithm, BSC provides additional benefits: (a)  simplifies the computationally expensive M-step of traditional EM method; (b) applicable to mixed data types and (c) computational complexity is linear in the data points. This paper extends the BSC algorithm to perform efficient clustering w.r.t. the proposed image model.

Image segmentation based on region merging is one of the oldest techniques used in computer vision \cite{murphy2012machine}. Numerous existing methods which merge regions in a RGB image exploit color and edge information \cite{tremeau2000regions, nock2004statistical, peng2011automatic, martinez2013unsupervised}. For indoor scenes, the use of color is often unreliable due to numerous effects caused by spatially varying illumination \cite{gupta2013perceptual} and the presence of shadows. Therefore, for  indoor scenes, color based merging is not as effective as it is for outdoor scenes. On the other hand, for indoor scenes the planar surfaces are considered as important geometric primitives. They are often employed for scene decomposition \cite{silberman2012indoor, rusu2013semantic, gupta2013perceptual} and grouping coplanar segments into extended regions \cite{taylor2011segmentation}. This motivates us to develop a region merging algorithm by mainly exploiting planar property of regions instead of color. Recent work \cite{hasnat2014wmm} has shown that the concentration parameter ($\kappa$) of the directional distributions can be exploited for characterizing planar surfaces. We take this into account and efficiently exploit the concentration ($\kappa$) of the surface normals in order to accept or reject a merging operation.

This paper proposes a novel unsupervised RGB-D segmentation method. It begins by applying a joint clustering method on RGB-D image features (color, position and normals), which generates a set of regions. Next, it applies a statistical region merging method on resulting regions to obtain the final segmentation. We evaluate the proposed method using RGB-D images of the NYU depth database (NYUD2) \cite{silberman2012indoor} and compare our results with the state of the art unsupervised techniques. To benchmark the segmentation task, we consider commonly used evaluation metrics such as \cite{arbelaez2011contour, freixenet2002yet}: segmentation covering, probability rand index, variation of information, boundary displacement error and boundary based F-measure. Moreover, we also consider the computation time of comparable methods as a measure of evaluation. 

Finally, the contributions related to the work described in this paper can be highlighted as follows:
\begin{itemize}
\item A statistical RGB-D image generation model (section \ref{sssec:image_model}) that incorporates both color and geometric properties of the scene.
\item An efficient probabilistic joint clustering method (section \ref{ssec:jcsa_method}) which exploits the Bregman divergence \cite{banerjee2005clusteringb, boissonnat2010bregman, liu2012shape}. The method has the following properties: (a) performs clustering with respect to the proposed image model; (b) provides an intrinsic view of the indoor scene and (c) provides statistics w.r.t. the planar property of the regions.
\item A statistical region merging method (Section \ref{ssec:region_merging}) which satisfies certain region merging predicates. This method can be incorporated independently with any other existing indoor RGB-D scene segmentation method.
\item
A benchmark (Section \ref{ssec:experiments_results}) on the NYUD2 \cite{silberman2012indoor} for unsupervised scene segmentation. Results from the proposed method show that it is comparable w.r.t. the state of the art and better in terms of computational time. 
\end{itemize}

This work exploits our earlier work on image analysis using directional distribution \cite{hasnat2015mbhcfmm, hasnat2014wmm, hasnat2013hierarchical}. Moreover, it provides an extension of our recent work on RGB-D segmentation \cite{hasnat2014unsupervised} by including additional details and newer contributions, such as:
\begin{itemize}
\item We introduce\footnote{In order to cluster surface normal, we proposed two methods: one based on the Fisher distribution in \cite{hasnat2015mbhcfmm} and another based on the Watson distribution in \cite{hasnat2014wmm}. In this paper, we exploit both of them within a common framework.} a general framework which exploits: (a) two theoretical models based on directional statistics in $3D$ (Fisher and Watson distributions) and (b) information geometry (Bregman divergence).
\item We propose a general methodology that can be used with unambiguous direction (using Fisher distribution) or with ambiguous direction (using Watson distribution). 
\item For the study of indoor scenes (NYUD2 dataset \cite{silberman2012indoor}), the ambiguity\footnote{In our previous work \cite{hasnat2014unsupervised}, we used the toolbox of \cite{silberman2012indoor} which produced ambiguity \cite{rusu2013semantic} in the direction of the normals. In this paper, we decided to use the toolbox of \cite{ren2012rgb} which removes such ambiguity.} in surface normal is removed, which allows the use of Fisher distribution. New experimental results are discussed through a common framework based on both Fisher and Watson distributions.
\item Several additional image models are explored, discussed and corresponding results are provided. 
\item An enhanced discussion based on new experiments, additional illustrations and clarifications.
\end{itemize}

The outline of the rest of this paper is as follows: Section \ref{sec:background_rgbd} discusses the background of RGB-D segmentation methods and related works. Section \ref{sec:methodology} presents the proposed method. Section \ref{sec:results_discussion} provides experimental results and discussion. Finally, Section \ref{sec:conclusion} draws conclusions and discusses future perspectives.

\section{Background of RGB-D Segmentation}
\label{sec:background_rgbd}
Color image segmentation of natural and outdoor scene is a well-studied problem due to its numerous applications in computer vision. 
Different methods have been already proposed in the state of the art based on different perspectives.
Chapter 5 of \cite{szeliski2011computer} provides a detail overview of these methods.

Many of the established image analysis methods have been either extended or directly employed to the depth image data in order to deal with depth features, see Chapter 6 of \cite{dal2012time} for a detail review. In the simplest cases, the depth image is considered as a grayscale image or converted to a cloud of 3D points. However, such simple approaches have limitations \cite{dal2012time}. 
For example, clustering using only 3D points often fails to locate the intersections among planar surfaces with different orientations such as wall, floor, ceiling, etc.  This is due to the fact that the 3D points associated to the intersections are grouped into a single cluster.
For this reason, better features such as surface normals are suggested to use \cite{rusu2013semantic, holz2012real}. However, from a recent study \cite{hasnat2014wmm}, we observe that: (a) the use of surface normals solely is not sufficient to extract full semantics of the scene, e.g., multiple objects with nearly similar orientations may grouped into the same cluster irrespective of their 3D location and (b) it is necessary to incorporate additional features, such as color, texture, etc. to provide better interpretation of indoor environments. Such observations raise the necessity to jointly exploit depth, color and other features for the task of RGB-D image analysis.

A number of recent research activities, such as \cite{ dal2012fusion}, \cite{gupta2013perceptual}, \cite{ren2012rgb} and \cite{silberman2012indoor}, proposed different methodologies for indoor scene understanding and analysis with promising results. Most of these researches incorporate depth as complementary information with color images. They can be categorized mainly from two aspects: (a) \textit{feature-wise}: different types, levels and dimensions of features and (b) \textit{method-wise}: numerous distinctions, such as supervised, unsupervised, clustering based, graph based, split-merge based, etc. Different methods emphasize on different aspects of the problem, which in general opens a number of interesting and challenging issues to focus on.

A common approach to tackle the RGB-D scene analysis problem is to extract different features, design kernels and classify pixels with learned classifiers. For example, \cite{ren2012rgb} proposed contextual models in a supervised setting. Their model combines kernel descriptors with a segmentation tree or with superpixels Markov Random Field (MRF). To this aim, they extended the well-known gPb-UCM algorithm \cite{arbelaez2011contour} to incorporate the global probability of boundaries (gPb) of depth image with gPb of RGB image. The RGB-D scene analysis method proposed by \cite{silberman2012indoor} first gives an over-segmentation of the scene by applying watershed on the gPb of the RGB image. Next, it aligns the over-segmentation with the 3D planes. Finally, using a trained classifier it applies a hierarchical segmentation in order to merge regions. Another interesting feature of \cite{silberman2012indoor} is that it provides an annotated RGB-D dataset (NYUD2) to perform scene analysis. Recently, \cite{gupta2013perceptual} extended the gPb-UCM \cite{arbelaez2011contour} method to a supervised setting. First, they combine geometric contour cues: convex and concave normal gradients with monocular cues: brightness, color, texture. Then, they detect pixels as contours via learned classifiers for 8 different orientations. Finally, they generate a hierarchy of segmentations from all oriented detectors. All of the above-mentioned methods use supervised approach in order to combine/fuse different features or information extracted from them. Let us now focus on methods developed for the unsupervised domain.

\cite{dal2012fusion} discussed about the fusion of color with geometry based on an unsupervised setting and provide a solution using the normalized cut spectral clustering method. Their approach consists of identifying an optimal multiplier to balance between color and depth. For this reason, they generate several segmentations with different values of the multiplier. Each segmentation is obtained by applying spectral clustering on the fused subsampled features. Finally, they select the best segmentation based on their proposed RGB-D segmentation quality evaluation score. In practice, this method requires more computation time than others as it generates a number of different segmentations for a single image. \cite{taylor2011segmentation} proposed a method which first extracts edges from RGB image, applies Delaunay Triangulation on edges to construct triangular graph and then applies Normalized Cut algorithm to the graph. In a second step, they extract planar surfaces from the segments using RANSAC \cite{szeliski2011computer} and finally merge the coplanar segments using a greedy merging procedure. The unsupervised method that we propose in this paper is different than the above proposals as: (a) it considers surface normals as features; (b) it employs mixture model based joint clustering rather than Normalized Cut and (c) it merges regions based on statistics rather than a greedy approach. 

Beside these approaches, the well-known graph based segmentation \cite{felzenszwalb2004efficient} is extended for joint color and depth image segmentation. For example, \cite{niu2012leveraging} extended it by including disparity with color for the purpose of segmenting stereopsis images. \cite{strom2010graph} extended it by incorporating surface normals to segment colored 3D laser point clouds. For the purpose of comparison, we develop an extension of the graph based method that considers both 3D and normals along with color.

Despite all of these researches, it remains an interesting issue about how to build an appropriate statistical model to describe RGB-D images of indoor scenes and how to exploit such model to segment the captured images. Scene-SIRFS \cite{barron2013intrinsic} is a recently proposed model whose aim is to recover intrinsic scene properties from single RGB-D image. It considers a mixture of shapes and illuminations where the mixture components are embedded in a soft segmentation of 17 eigenvectors. These eigenvectors are obtained from the normalized Laplacian corresponding to the input RGB image. Although the concept of using mixture is similar to the method proposed in this paper, the underlying objective, model and methodologies are different. We consider a mixture of shape (via 3D and normals) and color that consists of a feature vector of length 9. In the next Section, we present our proposed scene analysis method.
\section{Methodology}
\label{sec:methodology}
In this section, we present the proposed RGB-D segmentation method. First, we discuss the statistical image generation model and present the segmentation method w.r.t. the model. Then we briefly present the joint clustering method followed by the region merging method.

\subsection{Model and method}
\label{ssec:model_method}
\subsubsection{Image Generation Model}
\label{sssec:image_model}
\begin{figure}[H]
\centering
\includegraphics[scale=0.38]{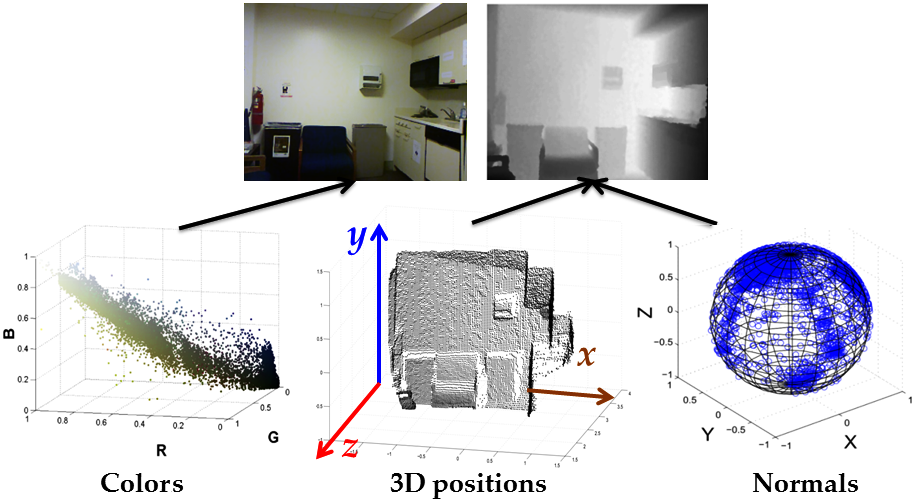}
\caption{Illustration of the proposed image generation model. The first row shows the color and depth image. The second row shows the features of the model in their respective spaces.}
\label{fig:image_model}
\end{figure}

We propose a statistical image model that fuses color and shape (3D and surface normals) features according to the \emph{“na\"{\i}ve” Bayes} assumption \cite{zhang2004optimality, lewis1998naive}, i.e., the features are independent of each other. Furthermore, it is based on a generative model-based approach \cite{murphy2012machine}, where the features are issued from a finite mixture of different probability distributions. Figure \ref{fig:image_model} provides an illustration of the proposed image generation model. We can observe that, the color and 3D features belong to the standard Euclidean space, i.e., in $\mathbb{R}^3$ and the surface normal\footnote{Surface normal is a 3D unit vector that describes the planar property of a pixel. This planar property is the perpendicular direction to the plane which is fitted on each pixel using chosen neighboring pixels. In the unit sphere of Figure \ref{fig:image_model}, each blue point indicates the direction of a pixel's normal w.r.t. the origin of the sphere.} belongs to the unit sphere, i.e., in $S^{2}$. Based on this observation, we consider the multivariate Gaussian \cite{bishop2006pattern} distribution for the color and 3D features and the directional\footnote{We use the term \textit{directional} for the \textit{Fisher} and the \textit{Watson} distribution. Both of them are parameterized with a mean direction $\mu$ and concentration value $\kappa$. They belong to the regular exponential family, which allows us to provide a common formulation despite their different normalization function, see Section \ref{sssec:mvmfmm_dist} and \ref{sssec:mwmm_dist}.} (Fisher or Watson) \cite{mardia2009directional, hasnat2015mbhcfmm, hasnat2014wmm} distribution for surface normals. Mathematically, such a model with $k$ components has the following form:
\begin{equation}
\begin{multlined}
\label{eq:ggw_mm}
g\left ( \mathbf{x}_{i}|\Theta_{k} \right ) = \sum_{j=1}^{k}\pi_{j,k} \, f_g(\mathbf{x}_i^C|\mu_{j,k}^C, \Sigma_{j,k}^C) \, f_g(\mathbf{x}_i^P|\mu_{j,k}^P, \Sigma_{j,k}^P) \\
f_{dir}\left ( \mathbf{x}_i^N|\mu_{j,k}^N,\kappa_{j,k}^N\right )
\end{multlined}
\end{equation}
Here $\mathbf{x}_{i} = \{ \mathbf{x}_{i}^C, \mathbf{x}_{i}^P, \mathbf{x}_{i}^N \}$ is the 9 dimensional feature vector of the $ith$ pixel with $i=1,...,M$. Superscripts denote: C - color, P - 3D position and N - normal. $\Theta_k = \{\pi_{j,k}, \AB \mu_{j,k}^C, \AB \Sigma_{j,k}^C, \AB \mu_{j,k}^P, \AB \Sigma_{j,k}^P, \AB \mu_{j,k}^N, \AB \kappa_{j,k}^N\}_{j=1...k}$ denotes the set of model parameters where $\pi_{j,k}$ is the prior probability, $\mu_{j,k} = \lbrace \mu_{j,k}^C, \mu_{j,k}^P, \mu_{j,k}^N\rbrace$ is the mean, $\Sigma_{j,k} = \lbrace \Sigma_{j,k}^C, \Sigma_{j,k}^P\rbrace$ is the variance-covariance symmetric positive-definite matrix and $\kappa_{j,k}^N$ is the concentration of the $jth$ component. $f_g(.)$ and $f_{dir}(.)$ are the density functions of the multivariate Gaussian distribution (Section \ref{sssec:mgmm_dist}) and the directional (Fisher or Watson) distribution (Section \ref{sssec:mvmfmm_dist} and \ref{sssec:mwmm_dist}) respectively.

\subsubsection{Segmentation method}
\label{sssec:method_segmentation}
\begin{figure*}[!t]
\centering
\subfloat[]{\includegraphics[scale=0.9]{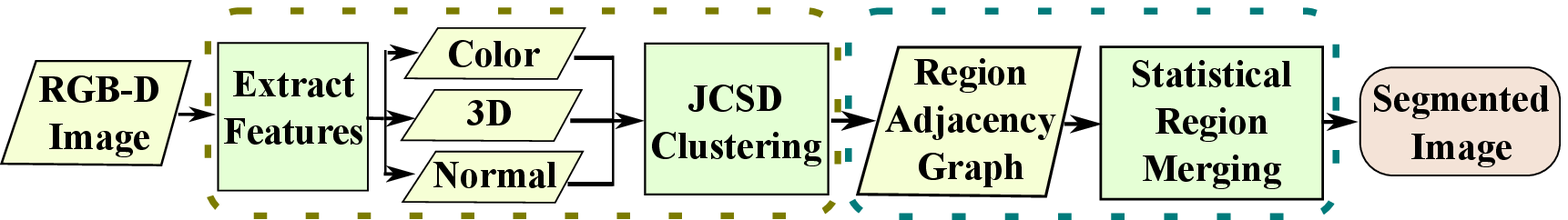}}
\hfil
\subfloat[]{\includegraphics[scale=0.67]{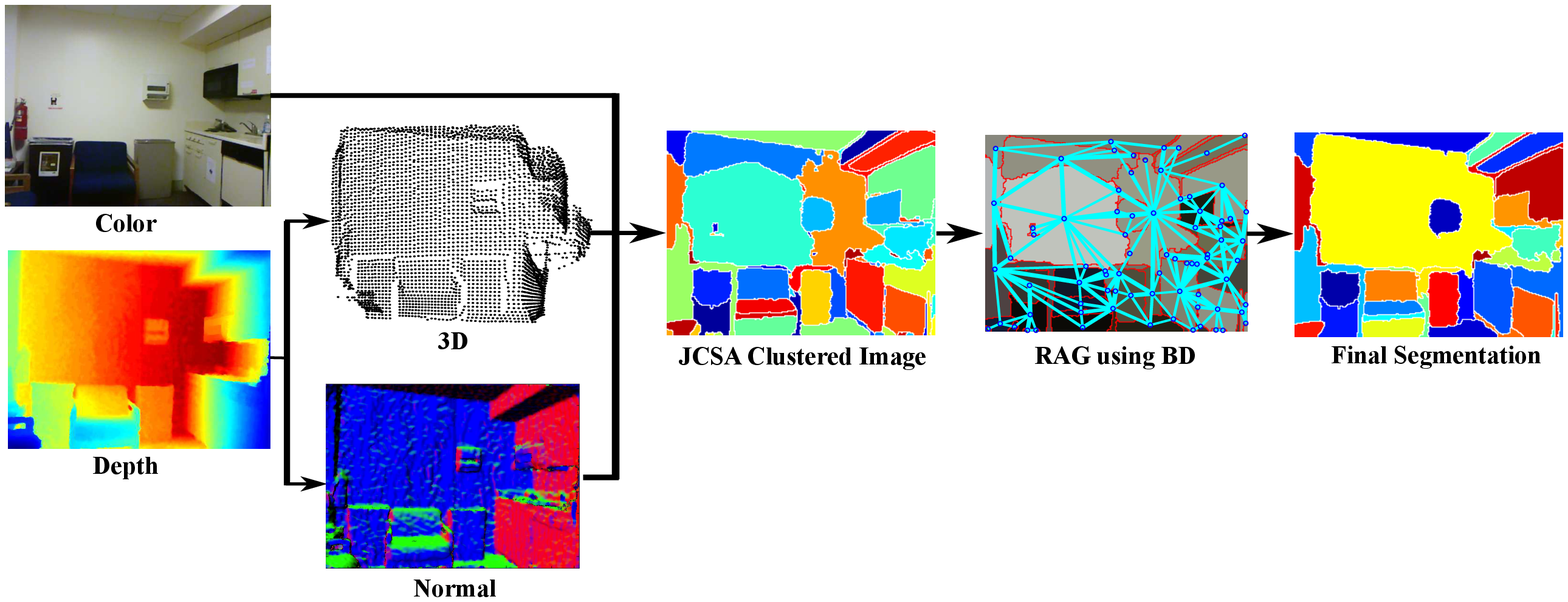}}
\caption{Work flow of the proposed segmentation method. (a) Block diagram and (b) Illustration with an example.}
\label{fig:block_diagram}
\end{figure*}

Figure \ref{fig:block_diagram} illustrates the workflow of the proposed RGB-D segmentation method that consists of two sub-tasks: (1) clustering heterogeneous (color, 3D and Normal) data and (2) merging regions. The first task performs a joint color-spatial-directional clustering and generates a set of regions. The second task performs a refinement on this set with the aim to merge regions which are susceptible to be over-segmented. In the next two sub-sections we present our methods to accomplish these tasks.

\subsection{Joint Color-Spatial-Directional (JCSD) clustering}
\label{ssec:jcsa_method}
In order to cluster heterogeneous data, we develop a Joint Color-Spatial-Directional (JCSD) clustering method. The clustering method estimates the parameters of the mixture model (Eq. (\ref{eq:ggw_mm})) as well as clusters the image data/features. As an outcome, we obtain the groups of image pixels which form the regions in the image. However, notice that in an unsupervised setting the true number of segments are unknown.
Therefore, we cluster features with the assumption of a known maximum number of clusters ($k=k_{max}$). Section \ref{ssec:discussion} provides additional details on this issue. Such assumption often causes an over-segmentation of the image. In order to overcome this issue, it is necessary to merge the over-segmented regions (see Section \ref{ssec:region_merging}).

The proposed joint clustering method follows the Bregman Soft Clustering (BSC) method \cite{banerjee2005clusteringb} and extends it to combine multiple probability distributions which belong to the REF.
The extension is based on the independence assumption to combine different distributions for different types of features.
This allows computing the divergence among two distributions based on the following combined form: 
\begin{equation}
\begin{multlined}
\label{eq:comb_prob_form}
f_{comb}(\mathbf{x}_i|\Theta_{j,k}) = f_g(\mathbf{x}_i^C|\mu_{j,k}^C, \Sigma_{j,k}^C) \, f_g(\mathbf{x}_i^P|\mu_{j,k}^P, \Sigma_{j,k}^P) \\
 f_{dir} \left ( \mathbf{x}_i^N|\mu_{j,k}^N,\kappa_{j,k}^N\right )
\end{multlined}
\end{equation}
where $\Theta_{j,k}= \{\pi_{j,k}, \mu_{j,k}^C, \Sigma_{j,k}^C, \mu_{j,k}^P, \Sigma_{j,k}^P, \mu_{j,k}^N,\kappa_{j,k}^N\}$ denotes the $j_{th}$ component of parameter $\Theta_k$. This allows to develop a joint Bregman soft clustering method for the model in Eq. (\ref{eq:ggw_mm}). 
\subsubsection{Regular Exponential Family (REF) of Distributions and Bregman Divergence}
\label{sssec:efd}
A multivariate probability density function $f(x|\eta)$ belongs to the Regular Exponential Family (REF) \cite{liu2012shape} if it has the following (see Eq. (3.7) of \cite{banerjee2005clusteringb}, Eq. (60) of \cite{nielsen2009statistical}) form\footnote{In order to keep our formulations concise, we use the expectation parameters $\eta$ to define the REF distributions. However, the other form (see Appendix \ref{ssec:bd_efd}) and related derivations are available in \cite{garcia2010simplification} (for the Gaussian distribution) and \cite{hasnat2015mbhcfmm, hasnat2014wmm} (for the Fisher and Watson distributions).}:
\begin{equation} 
\label{eq:efd}
f\left(x|\eta\right ) = \text{exp}\left (-D_G\left( t(x), \eta \right)\right )\text{exp} \left(k(x)\right)
\end{equation} 
and
\begin{equation}
\label{eq:BD_eta}
D_{G}\left ( \eta_1, \eta_2 \right )= G(\eta_1) - G(\eta_2) - \left \langle \eta_1 - \eta_2, \nabla G(\eta_2) \right \rangle
\end{equation}
with $G(.)$ is the Legendre dual of $F(.)$. $F(.)$ is a strictly convex log normalizing function associated with a probability distribution. $\nabla G$ is the gradient of $G$. $t(x)$ denotes the sufficient statistics and $k(x)$ is the carrier measure. The expectation of the sufficient statistics $t(x)$ w.r.t. the density function (Eq. (\ref{eq:efd})) is called the expectation parameter $(\eta)$. 
$D_{G}$ is the Bregman Divergence (BD) \cite{banerjee2005clusteringb, boissonnat2010bregman, liu2012shape} computed from expectation parameters, see Appendix \ref{ssec:bd_efd}. BD can be used as a measure of dissimilarity between two distributions of the same exponential family which are defined by two expectation parameters $\eta_1$ and $\eta_2$. We will define in the following Section the particular forms obtained with the Gaussian distribution and the directional (Fisher and Watson) distribution.
\subsubsection{Multivariate Gaussian Distribution}
\label{sssec:mgmm_dist}
For a $d$ dimensional random vector $\mathbf{x} = \left[x_1,...,x_d \right]^T \in \R^d$, the multivariate Gaussian distribution is defined as:
\begin{equation}
\begin{multlined}
\label{eq:mgmm}
f_{g}(\mathbf{x}|\mu, \Sigma)= \\
\frac{1}{(2\pi)^{d/2} \;\text{det}(\Sigma)^{1/2}}\,\text{exp}\left (- \frac{1}{2}(\mathbf{x}-\mu)^T \, \Sigma^{-1} \, (\mathbf{x}-\mu) \right)
\end{multlined}
\end{equation} 
Here, $\mu \in \R^d$ denotes the mean and $\Sigma$ denotes the variance-covariance symmetric positive-definite matrix. To write the multivariate Gaussian distribution in the form of Eq. (\ref{eq:efd}), the elements are defined as \cite{nielsen2009statistical}: sufficient statistics $t(\mathbf{x})= (\mathbf{x}, -\mathbf{x}\mathbf{x}^T)$; carrier measure $k(\mathbf{x})=0$; expectation parameter $\eta = (\phi, \Phi) = \left(\mu, -(\Sigma + \mu \mu^T)\right)$ and 
\begin{equation}
\label{eq:dlnf_gmmd}
G_g(\eta) = -\frac{1}{2} \;\text{log}(1 + \phi^T \Phi^{-1} \phi) - \frac{1}{2} \text{log}(\text{det}(\Phi)) - \frac{d}{2}\text{log}(2\pi e)
\end{equation}
\subsubsection{Fisher Distribution}
\label{sssec:mvmfmm_dist}
For a $3$-dimensional random unit vector $\mathbf{x} = \left[x_1,x_2,x_3 \right]^T \in S^{2} \subset \R^3$ (i.e., $\left \| \mathbf{x} \right \|_2=1$), the Fisher distribution is defined as \cite{mardia2009directional, hasnat2015mbhcfmm}:
\begin{equation}
\label{eq:vmf_d_dim}
f_{dir}(x|\mu, \kappa)= \frac{\kappa}{\text{sinh}(\kappa)}\;\text{exp}(\kappa \mu^{T}x)
\end{equation} 
Here, $\mu$ denotes the mean (with $\left \| \mu \right \|_2=1$) and $\kappa$ denotes the concentration parameter (with $\kappa\geq 0$). The Fisher distribution is a special case of the von Mises-Fisher (vMF) \cite{mardia2009directional} distribution for three dimensional observations.
To write the Fisher distribution in the form of Eq. (\ref{eq:efd}), the elements are defined as \cite{hasnat2013hierarchical, hasnat2015mbhcfmm}: sufficient statistics $t(x)= x$; carrier measure $k(x)=0$; expectation parameter $\eta = \left \| \eta \right \|_2 \mu$  and 
\begin{equation}
\label{eq:dlnf_vmfd}
G_{dir}(\eta) = \kappa \left \| \eta \right \|_2 - \text{log} \left(\frac{\kappa}{\text{sinh}(\kappa)} \right)
\end{equation}
With the above formulation, for a set of observations $\mathbf{X} = \{ \mathbf{x}_{i} \}_{i=1,...,M}$ we estimate $\eta = E[t(\mathbf{X})]$ and $\kappa$ with a Newton-Raphson root finder method as \cite{hasnat2013hierarchical, hasnat2015mbhcfmm}:
\begin{equation}
\label{eq:NR_solve_vmf}
\kappa_{l+1} = \kappa_{l} - \frac{a-b-\left \| \eta \right \|_2}{1-a^{2}+b^{2}}
\end{equation}
where, $a=\text{tanh}(\kappa)^{-1}$ and $b=(\kappa)^{-1}$.
\subsubsection{Multivariate Watson Distribution}
\label{sssec:mwmm_dist}
For a $d$ dimensional unit vector $\mathbf{x} = \left[x_1,...,x_d \right]^T \in S^{d-1} \subset \R^d$ (i.e. $\left \| \mathbf{x} \right \|_2=1$), the multivariate (axially symmetric, i.e., $f_{dir}(\mathbf{x}|\mu,\kappa) = f_{dir}(\mathbf{-x}|\mu,\kappa) $) Watson distribution (mWD) is defined as \cite{mardia2009directional}:
\begin{equation}
\label{eq:mw_d_dim}
f_{dir}(\mathbf{x}|\mu,\kappa) = M \left( 1/2,d/2,\kappa \right)^{-1} \text{exp}\left(\kappa(\mu^{T} \mathbf{x})^2 \right)
\end{equation} 
Here, $\mu$ is the mean direction (with $\left \| \mu \right \|_2=1$), $\kappa \in \R$ the concentration and $M \left(1/2,d/2,\kappa \right)$ the Kummer's function \cite{mardia2009directional}. To write the mWD in the form of Eq. (\ref{eq:efd}), the elements are defined as \cite{hasnat2014wmm}: sufficient statistics $t(\mathbf{x})= \left[x_1^2,...,x_d^2, \sqrt{2}x_1 x_2,...,\sqrt{2}x_{d-1} x_d \right]^T$; carrier measure $k(\mathbf{x})=0$; expectation parameter $\eta$: 
\begin{equation}
\label{eq:exp_parm_wd}
\eta = \left \| \eta \right \|_2 \nu
\end{equation}
where $\nu=\left[\mu_1^2,...,\mu_d^2, \sqrt{2}\mu_1 \mu_2,...,\sqrt{2}\mu_{d-1} \mu_d \right]^T$ and
\begin{equation}
\label{eq:dlnf_wd}
G_{dir}(\eta) = \kappa \left \| \eta \right \|_2 - \text{log} M \left(1/2,d/2,\kappa \right)
\end{equation}
With the above formulation, for a set of observations $\mathbf{X} = \{ \mathbf{x}_{i} \}_{i=1,...,M}$ we estimate $\eta = E[t(\mathbf{X})]$ and $\kappa$ with a Newton-Raphson root finder method as \cite{hasnat2014wmm}:
\begin{equation}
\label{eq:NR_solve_w}
\kappa_{l+1}= \kappa_{l} - \frac{ q(1/2,d/2;\kappa_{l})- \left \| \eta \right \|_2 }{q^{'}(1/2,d/2;\kappa_{l})}
\end{equation}
where $q(1/2,d/2;.)$ is the Kummer-ratio, $q^{'}(1/2,d/2;.)$ is the derivative of $q(1/2,d/2;.)$.
\subsubsection{Bregman Divergence for the combined model}
\label{sssec:bd_combined}
Our image model (in Eq. (\ref{eq:ggw_mm})) combines different exponential family of distributions (associated to color, 3D and normals) based on independent (\emph{na\"{\i}ve Bayes} \cite{zhang2004optimality, lewis1998naive}) assumption. Therefore, Bregman Divergence (BD) \cite{banerjee2005clusteringb, boissonnat2010bregman, liu2012shape} of the combined model can be defined as a linear combination of the BD of each individual distributions:
\begin{equation}
\label{eq:BD_comb}
D_{G}^{comb}(\eta_{i},\eta_{j})= D_{G,g}^{C}(\eta_{i}^C,\eta_{j}^C) + D_{G,g}^{P}(\eta_{i}^P,\eta_{j}^P) + D_{G,dir}^{N}(\eta_{i}^N,\eta_{j}^N)
\end{equation}
where, $D_{G,g}(.,.)$ denotes BD using the multivariate Gaussian distribution \cite{garcia2010simplification} and $D_{G,dir}(.,.)$ denotes BD using the directional (Fisher or Watson) distribution \cite{hasnat2015mbhcfmm}. Then, it is possible to define, with expectation parameter $\eta=\left\{\eta^C,\eta^P,\eta^N\right\}$:
\begin{equation}
\label{eq:G_comb}
G^{comb}(\eta)= G_g(\eta^{C}) + G_g(\eta^{P}) + G_{dir}(\eta^{N})
\end{equation}
\subsubsection{Bregman Soft Clustering for the combined model}
\label{sssec:bsc_combined}
Bregman Soft Clustering (BSC) exploits Bregman Divergence (BD) in the Expectation Maximization (EM) \cite{mclachlan2008em} framework to compute the Maximum Likelihood Estimate (MLE) of the mixture model parameters and provides a soft clustering of the observations \cite{banerjee2005clusteringb}. 

In order to cluster data with the combined model (Eq. (\ref{eq:ggw_mm})), it is necessary to estimate the model parameters and obtain $\hat{\Theta}_k$ for $g(\mathbf{X}|\Theta_k)$ such that:
\begin{equation}
\label{eq:wmm_opt_params}
\hat{\Theta}_k=\operatorname*{arg\,max}_{\Theta_k} g\left ( \mathbf{X} |\Theta_{k} \right ) \text{with } g(\mathbf{X} |\Theta_k)= \prod_{i=1}^M g(\mathbf{x}_i |\Theta_k ) 
\end{equation}
Here, $\mathbf{X}= \left \{\mathbf{x}_i \right \}_{i=1,...,M}$ is the set of observations. Let $\gamma_i=j$ denotes the class label of an observation $\mathbf{x}_i$ with $j=\lbrace1,\ldots,k\rbrace$.

BSC consists of an Expectation step (E-step) and a Maximization step (M-step). In the E-step of the algorithm, the posterior probability is computed as \cite{nielsen2009statistical}:
\begin{equation}
\begin{multlined}
\label{eq:posterior_bsc}
p\left ( \gamma_i=j|\mathbf{x}_{i} \right ) = \\
\frac{\pi_{j,k}\,\text{exp}\left ( G^{comb}(\eta_{j,k}) + \left \langle t(\mathbf{x}_{i}) - \eta_{j,k}, \nabla G^{comb}(\eta_{j,k}) \right \rangle \right )}{\sum_{l=1}^{k}\pi_{l,k}\,\text{exp}\left ( G^{comb}(\eta_{l,k}) + \left \langle t(\mathbf{x}_{i}) - \eta_{l,k}, \nabla G^{comb}(\eta_{l,k}) \right \rangle \right )}
\end{multlined}
\end{equation}
Here, $\eta_{j,k}$ and $\eta_{l,k}$ denote the expectation parameters for any cluster $j$ and $l$ given that the total number of components is $k$. The M-step updates the mixing proportion and expectation parameter for each class as:
\begin{equation}
\label{eq:maximization_bsc}
\pi_{j,k}=\frac{1}{M}\sum_{i=1}^{M}p\left ( \gamma_i=j|\mathbf{x}_{i} \right ) \,\, \text{and} \,\, \eta_{j,k}=\frac{\sum_{i=1}^{M}p\left ( \gamma_i=j|\mathbf{x}_{i} \right )\mathbf{x}_{i}}{\sum_{i=1}^{M}p\left ( \gamma_i=j|\mathbf{x}_{i} \right )}
\end{equation}

Initialization (using the EM method) is a challenging issue and has significant impact on clustering \cite{biernacki2003Choosing}. Our initialization procedure consists of setting initial values for prior class probability $(\pi_{j,k})$ and the expectation parameters $(\eta_{j,k})$ with $1\leq j\leq k$. 
We obtain these initial values for the Gaussian and directional (Fisher or Watson) distributions using a combined k-means type clustering. After initialization, we iteratively apply the E-step and M-step until the convergence criteria are met. These criteria are based on maximum number of iterations (e.g., 200) and a threshold difference (e.g., 0.001) between the negative log likelihood values (see Eq. (\ref{eq:n_llh}) and Eq. (\ref{eq:ggw_mm})) of two consecutive steps.
\begin{equation}
\label{eq:n_llh}
nLLH(\hat{\Theta}_k) = -\sum_{i=1}^{M}\text{log} \left ( g \left ( \mathbf{x}_{i}|\hat{\Theta}_k \right ) \right )
\end{equation}
The above procedures lead to a soft clustering, which generates associated probabilities and parameters for each component of the proposed model defined by Eq. (\ref{eq:ggw_mm}). Finally, for each sample we get the cluster label ($\hat{\gamma}_{i}$) using the updated combined BD (Eq. \ref{eq:BD_comb}) as:
\begin{equation}
\label{eq:hard_clust}
\hat{\gamma}_{i} = \operatorname*{arg\,min}_{j=1,...,k}D_G^{comb}(t(x_{i}),\hat{\eta}_{j,k})
\end{equation}
Applying Eq. (\ref{eq:hard_clust}) performs hard clustering on the data. Let us call this entire clustering method the BSC-COMB algorithm (Algorithm \ref{algo:bsc_comb}). This method can be seen as a general algorithm for combining different types of REF probability distributions with an independent assumption. However, to be more specific, in the experimental section we will denote it as the joint color-spatial-directional (JCSD) algorithm. 
\begin{algorithm}
 \SetAlgoLined
 \KwIn{$\mathbf{X}= \left \{\mathbf{x}_{i} \,\, | \,\, \mathbf{x}_{i} = \{ \mathbf{x}_{i}^C, \mathbf{x}_{i}^P, \mathbf{x}_{i}^N \} \,\wedge \, 1 \leqslant i \leqslant M \right \}$}
 \KwOut{Clustering of $\mathbf{X}$ with $k$ components.}
 Initialize $\pi_{j,k}$ and $\eta_{j,k}$ for $1\leq j\leq k$ using combined k-means\;
 \While{not converged}{
  \{Perform the E-step of EM\}\;
  \ForEach{ $i$ and $j$}{
  Compute $p(\gamma_i=j|\mathbf{x}_{i})$ using Eq. (\ref{eq:posterior_bsc})
  }
  \{Perform the M-step of EM\}\;
  \For{$j=1$ to $k$}{
  Update $\pi_{j,k}$ and $\eta_{j,k}$ using Eq. (\ref{eq:maximization_bsc})
  } 		
}
 Define final values of parameters as $\hat{\pi}_{j,k}$ and $\hat{\eta}_{j,k}$
 Assign each observation to a cluster using Eq. (\ref{eq:hard_clust})
 \caption{BSC-COMB (also called JCSD) algorithm for Joint Color-Spatial-Directional clustering.}
 \label{algo:bsc_comb}
\end{algorithm}

Applying Algorithm \ref{algo:bsc_comb} on RGB-D image features (color, position and normals) performs a joint color-spatial-directional clustering. This clustering method is based on the assumption of a known maximum number of components $k=k_{max}$. Image regions obtained by such clustering often lead to over-segmentation, see Figure \ref{fig:block_diagram}(b) for example. Therefore, it is necessary to merge the over-segmented regions. In the following section, we propose a region merging method to overcome such over-segmentation problem.
\subsection{Region Merging}
\label{ssec:region_merging}
In this step, we merge the over-segmented regions which are generated from previous step, i.e., after applying the JCSD clustering on the RGB-D image features. To this aim, first we build a Region Adjacency Graph (RAG) \cite{tremeau2000regions} (see Figure \ref{fig:block_diagram}). This graph is defined such as each region is a node and each node has edges with its adjacent nodes. In order to weight the edge connectivity among nodes, we consider a measure of statistical distance among two regions. Moreover, we weight the boundary strength among regions by a measure of their eligibility to merge. Similar to the standard region merging methods \cite{tremeau2000regions, nock2004statistical, peng2011automatic}, we develop an approach which depends on region merging predicate and merging order. As an outcome of region merging we obtain the final segmentation.
\subsubsection{Region Adjacency Graph (RAG)}
\label{sssec:rag_define}
\begin{figure}[!t]
  \centering
  \centerline{\includegraphics[viewport = 220 20 1100 700, clip = true, scale=0.27]{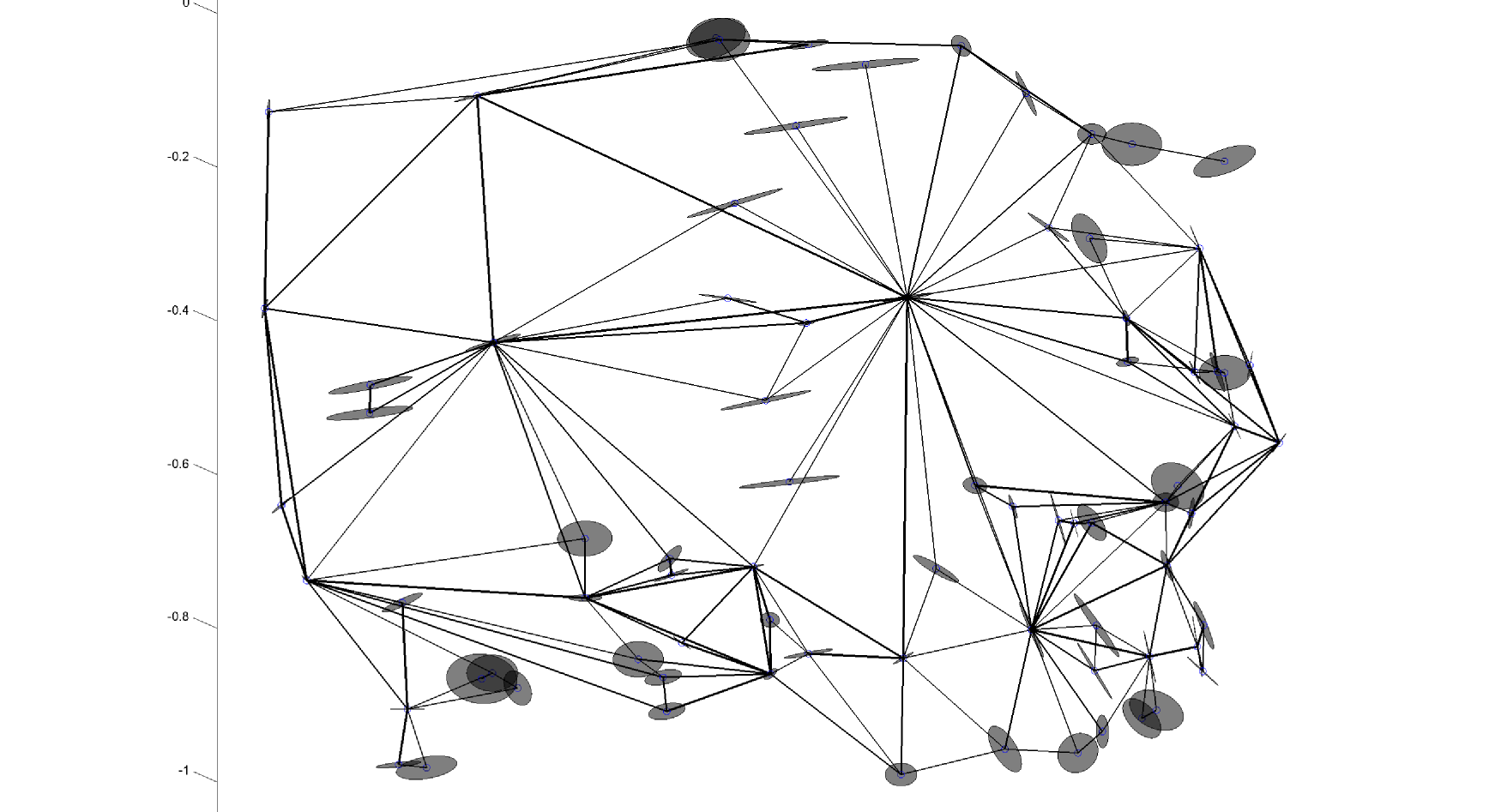}}
\caption{Illustration of a 3d view of the Region Adjacency Graph (RAG) constructed from JCSD clustered regions obtained from the image shown in Figure \ref{fig:block_diagram}(b). The circle associated to each node represents the concentration of image normals at the region. Each edge represents the weight $w_{d}$ associated to two adjacent regions. In this picture several circles resemble ellipses because of 3D to 2D projection. The 2D view of this graph overlaid on the original image is illustrated in Figure \ref{fig:block_diagram}(b).}
\label{fig:rag_image}
\end{figure}

In our proposed region merging method, RAG provides an inherent view of the merging strategy. From the JCSD clustered labels, we build the RAG by applying first a $3\mathbf{x}3$ median filter (in order to remove isolated and noisy labels) and then locating the regions from the enclosed boundaries. Figure \ref{fig:rag_image} illustrates an example of the RAG constructed from clustered regions obtained from the image shown in Figure \ref{fig:block_diagram}(b).
Let $R = \lbrace r_i \rbrace_{i=1,...,Z}$ be the set of regions, $G = (V, E)$ be the undirected graph that represents the RAG, where $v_{i} \in V$ is the set of nodes corresponding to the regions $r_{i} \in R$ and $E$ is the set of edges among adjacent nodes.

Each node $v_i$ is characterized by the source parameters (mean direction $\mu$ and concentration $\kappa$) of the directional (Fisher or Watson) distribution (Section \ref{sssec:mvmfmm_dist}) associated to region $r_i$. In Figure \ref{fig:rag_image} the radius of the circles (nodes) represents the $\kappa$ value and the orientation of the circles represents the mean direction $\mu$. Besides, in order to merge nodes efficiently, we compute the probability ($\pi$) and the expectation parameter ($\eta$) for each node. For a region $r_i$, $\pi_i$ is computed as the ratio of the number of region pixels w.r.t. total number of image pixels and $\eta_i^N$ is computed as the mean of the normals of the region. 

Each edge $e_{ij}$ consists of two weights: $w_d$, based on statistical dissimilarity and $w_b$, based on boundary strength between adjacent nodes $v_i$ and $v_j$.
The dissimilarity based weight $w_d$ is computed using the Bregman divergence (Eq. (\ref{eq:BD_eta})) among two adjacent nodes $v_i$ and $v_j$ as:
\begin{equation} 
\label{eq:wt_distance_regions}
w_d(v_i, v_j) = \text{min} \left ( D_{G,dir}^N(\eta_{i}^N, \eta_{j}^N), D_{G,dir}^N(\eta_{j}^N, \eta_{i}^N) \right ) 
\end{equation} 
where, $D_{G,dir}^N(\eta_{i}^N, \eta_{j}^N)$ is the Bregman divergence (Eq. (\ref{eq:BD_eta})) among the directional (Fisher or Watson) distributions associated with regions $r_i$ and $r_j$.
The boundary based weight $w_b$ between two nodes $v_i$ and $v_j$ is computed from the average normalized gradient values along the boundary of their corresponding regions $r_i$ and $r_j$ as:
\begin{equation} 
\label{eq:wt_avg_norm_grad}
w_b(v_i, v_j) = \frac{1}{\left | r_i\bigcap r_j \right |}\sum_{b\in r_i\bigcap r_j} I_G^{rgbd}(b)
\end{equation}
where, $r_i\bigcap r_j$ is the set of boundary pixels among two regions, $\left |. \right |$ denotes the cardinality and $I_G^{rgbd}$ is the normalized magnitude of image gradient\footnote{To compute image gradient $\Delta I = \left ( \frac{\partial I(x,y)}{\partial x}, \frac{\partial I(x,y)}{\partial y} \right )$, with $\frac{\partial I(x,y)}{\partial x} \approx \frac{I(x+1,y)-I(x-1,y)}{2}$ and $\frac{\partial I(x,y)}{\partial y} \approx \frac{I(x,y+1)-I(x,y-1)}{2}$, we used the 'sobel' operator in MATLAB implementation.\label{footnote_mog}} (MoG) \cite{szeliski2011computer} computed from the RGB-D image. 
$I_G^{rgbd}$ is obtained by first computing MoG for each color channels ($I_G^{r}$, $I_G^{g}$, $I_G^{b}$) and depth ($I_G^{d}$) individually, and then taking the maximum of those MoGs at each pixel.
\subsubsection{Merging Strategy}
\label{sssec:rag_predicate_order}
\begin{figure}[h]
\centering
\centerline{\includegraphics[scale=0.27]{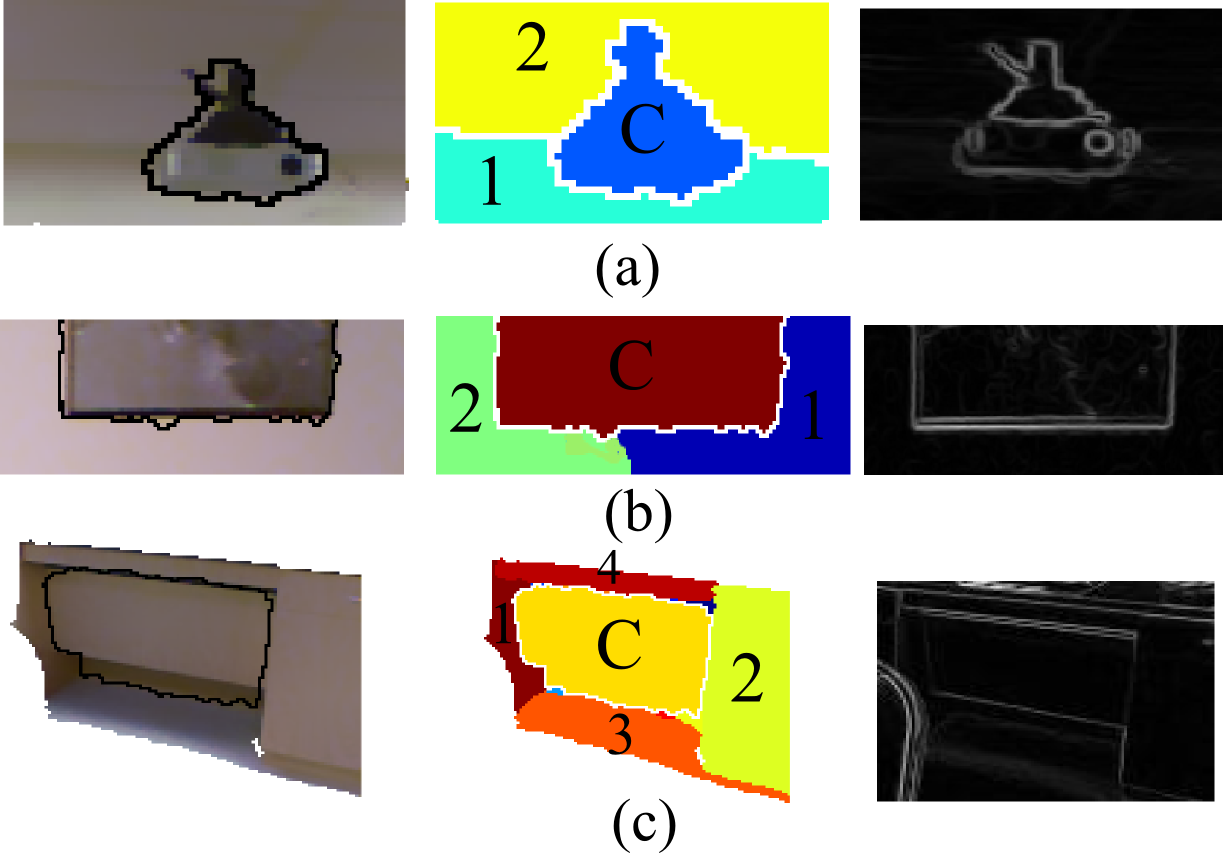}}
\caption{Illustration of the region merging predicate with different examples. The left column shows the RoI under process (surrounded by a black boundary) in the original image. The middle column shows the RoI under process (labeled as \textbf{C}) and the neighboring regions (labeled with numbers). The last column shows the magnitude of image gradient computed from the RGB-D image.}
\label{fig:rm_predicates}
\end{figure}

\begin{figure}[h]
\centering
\centerline{\includegraphics[scale=0.18]{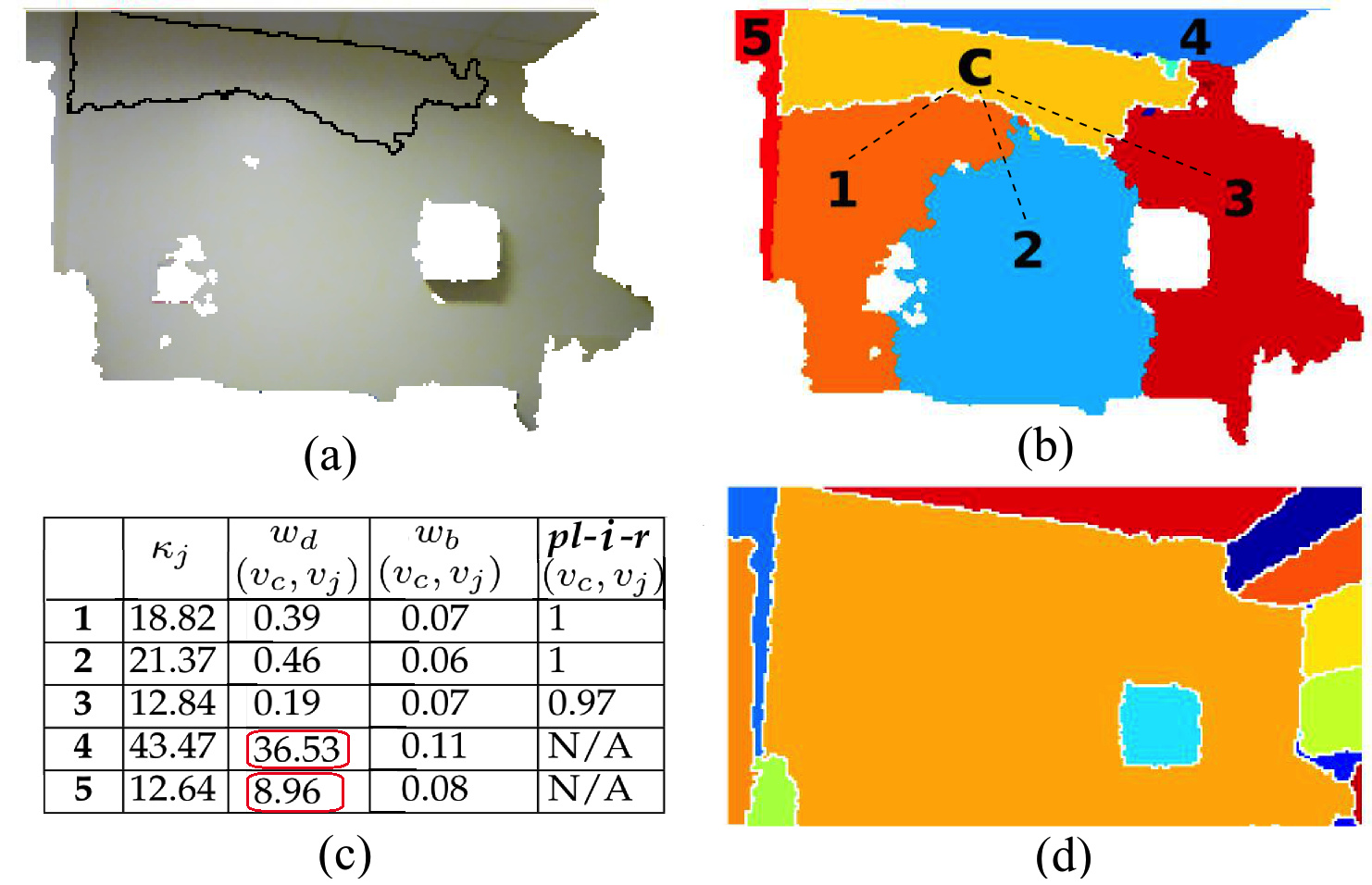}}
\caption{Illustration of the region merging strategy for a single region/node. (a) shows the RoI under process (surrounded by a black boundary) in the original image. (b) shows the RoI (labeled as \textbf{C}) and the neighboring regions (labeled with numbers). (c) provides the values (N/A means not necessary to compute) computed for the neighboring regions that could be merged with the RoI. (d) shows the merged regions after merging operation is completed for all RoIs.}
\label{fig:rm_example}
\end{figure}

Our region merging strategy is defined by an iterative procedure which is based on a merging predicate among adjacent nodes in a predefined order. The merging predicate consists of: (a) evaluating the \emph{candidacy} of each node; (b) evaluating the \emph{eligibility} of merging adjacent nodes and (c) verifying the \emph{consistency} of the merged nodes. Figure \ref{fig:rm_predicates} illustrates three examples to understand the merging predicate. Figure \ref{fig:rm_example} provides an example of the region merging strategy for a particular region/node. Once two nodes are merged, the information regarding the merged node and its edges are updated instantly. This procedure continues until no valid candidates are left to merge.

\emph{candidacy} of a node/region defines whether it is a valid candidate to be merged with the adjacent nodes. For each node, first we check its \emph{candidacy}. This helps us to filter out the nodes which are not valid candidates to be merged and hence reduces the computational time. For each node, our \emph{candidacy} criterion checks the planar property of the corresponding region. 
In indoor scenes either regions are planar (e.g. the floor, the walls, etc.) or are non planar (e.g. coffee pot, lamps, etc.). Whatever the method used, we noticed that most of over-segmentation errors are related to planar regions rather than non-planar regions (due to shadows, non-uniformity of lightness, etc.). Therefore we propose to use the planarity assumption as first criterion for region merging. As a consequence we propose to focus on adjacent planar regions and to avoid any region which is non planar. Indeed it makes more sense to merge two adjacent planar regions (if they have same depth and same color) than one non-planar region with its neighboring regions whatever the depth, color and planarity of these laters.
This planarity property can be easily investigated by analyzing the concentration parameter ($\kappa$) associated with each node $v_i$. We define the \textit{candidacy} of a node $v_i$ as follows:
\begin{equation} 
\label{eq:candidacy}
candidacy(v_i) = \begin{cases}
  true, & \text{if } \kappa_i > \kappa_{p}, \\
  false, & \text{otherwise}.
\end{cases}
\end{equation} 
Here $\kappa_i$ is the concentration parameter computed for the region $r_i$. $\kappa_{p}$ is the threshold that defines the planar property of a region. From our study on planar statistics, see Appendix \ref{apsec:study_planar_stat}, we observed that the concentration of the normals ($\kappa$) associated with a region can be exploited to discriminate among planar and non-planar surfaces. 
The Eq. (\ref{eq:candidacy}) was introduced to exploit this property.
See Section \ref{ssec:experiments_results} for details about this threshold value, which is set as $\kappa_p=5$. In Figure \ref{fig:rm_predicates}(a), the region/node of interest (labeled as \textbf{C}) has $\kappa_C=3$, which signifies that it is not a valid candidate for merging with the neighboring regions/nodes. Conversely, $\kappa_C=58$ in Figure \ref{fig:rm_predicates}(b) and $\kappa_C=11$ in Figure \ref{fig:rm_predicates}(c) means that those regions/nodes are valid candidates.
We define the \emph{eligibility} to merge two regions ($r_i$ and $r_j$) or nodes ($v_i$ and $v_j$) from the dissimilarity based weight $w_d$ (using Eq. (\ref{eq:wt_distance_regions})) and the boundary based weight $w_b$ (using Eq. (\ref{eq:wt_avg_norm_grad})) as:
\begin{equation} 
\label{eq:eligibility}
eligibility(v_i, v_j) = \begin{cases}
  true, &  (a) \,\, w_b(v_i, v_j) < th_b; \,\, \text{and} \\
        &  (b) \,\, w_d(v_i, v_j) < th_d; \\
  false, & \text{otherwise}.
\end{cases}
\end{equation} 
where, $th_b$ and $th_d$ are the thresholds associated with the boundary based weight $w_b$ (Eq. \ref{eq:wt_avg_norm_grad}) and the distance based weight $w_d$ (Eq. \ref{eq:wt_distance_regions}) respectively. See Section \ref{ssec:experiments_results} for the details of these threshold values, which are set as $th_d=3$ and $th_b = 0.2$. From our experiments on regions merging, we observed that most pairs of regions which have been selected to be merged have very low $w_d$. This motivates us to set a heuristic in order to verify the eligibility of merging two nodes $v_i$ and $v_j$ based on $w_d$. The use of the boundary/edge based weight $w_b$ is motivated from existing techniques such as the OWT-UCM \cite{arbelaez2011contour}.

In order to understand the impact of $w_b$ from an example, let us consider the regions in Figure \ref{fig:rm_predicates}(b), labeled as C, 1 and 2. All of them are valid candidates, because $\kappa_C=58$, $\kappa_1=81$ and $\kappa_2=53$. Boundary values are as follows: $w_b(v_C,v_1)=0.8$, $w_b(v_C,v_2)=0.7$ and $w_b(v_1,v_2)=0.03$, which signifies that the region C (region of interest) should not be merged with the neighboring regions 1 and 2. On the other hand, the regions 1 and 2 can be merged. Indeed, that makes more sense, from visual observation, to merge regions 1 and 2 (wall surfaces), rather than the region C (picture) with any of these two regions. As a consequence, the over-segmented walls, i.e., region 1 and 2, should be merged into a unique region. 

Now, in order to understand the impact of $w_d$ from an example, let us consider the two regions $v_1$ and $v_2$ of Figure \ref{fig:rm_predicates}(a), labeled as 1 and 2. They are valid candidates, because $\kappa_1=65$ and $\kappa_2=67$. Boundary value $w_b(v_1,v_2)=0.15$ means that they are eligible to merge. However, the dissimilarity value $w_d(v_1,v_2)=7$ is more than the threshold defined, which means that it does not make sense to merge regions 1 and 2 as the difference of their associated surface orientation is high. This is coherent with visual observation as the back wall (1) and the ceiling (2) should not be merged.

We employ the plane inlier ratio in order to verify the \emph{consistency} \cite{peng2011automatic} of a merged region. It is computed by first fitting a plane to the 3D points belonging to the merged region and then by computing the ratio of inliers and outliers based on a threshold distance \cite{taylor2013parsing}. We employed the widely used RANSAC \cite{szeliski2011computer} algorithm for the purpose of plane fitting. Therefore, we define \emph{consistency} among two regions $r_i$ and $r_j$ as follows:
\begin{equation} 
\label{eq:consistency}
consistency(v_i, v_j) = \begin{cases}
  true, & \text{if } pl\hbox{-}i\hbox{-}r (v_i, v_j)> th_r, \\
  false, & \text{otherwise}.
\end{cases}
\end{equation} 
where, $th_r$ is the threshold associated to the plane inlier ratio \textit{pl\hbox{-}i\hbox{-}r}. We set this threshold $th_r=0.9$ following the existing methods, such as \cite{taylor2013parsing}. We compute \textit{pl\hbox{-}i\hbox{-}r} by dividing the total number of inliers (3D points fitted within a plane based on a minimum/threshold distance) with the total number of 3D points used to fit the plane.

In order to understand the impact of $th_r$ from an example, let us consider the two regions $v_C$ and $v_2$ in Figure \ref{fig:rm_predicates}(c), labeled as C and 2. They are valid candidates, because $\kappa_C=11$ and $\kappa_2=15$. Boundary value $w_b(v_C,v_2)=0.13$ and dissimilarity value $w_d(v_C,v_2)=0.8$ means that they are eligible to merge. However, $pl\hbox{-}i\hbox{-}r(v_C,v_2)=0.84$ is less than the threshold, means that there is an inconsistency between regions C and 2 in terms of planar property when we try to merge them. This is coherent with visual observation as these two regions belong to two different planes localized at a different distance/depth.

Finally, we define the \emph{region merging predicate} \cite{peng2011automatic} $P_{ij}$ based on: (a) candidacy (using Eq. (\ref{eq:candidacy})); (b) eligibility of merging (using Eq. (\ref{eq:eligibility})) and (c) consistency of merged node (using Eq. (\ref{eq:consistency})) as:
\begin{equation} 
\label{eq:reg_merge_pred}
P_{ij} = \begin{cases}
  true, & \text{if } (a) \,\, candidacy(v_j) = true; \,\, \text{and} \\
        &  \,\,\,\,\,\, (b) \,\, eligibility(v_i, v_j) = true; \,\, \text{and} \\
        &  \,\,\,\,\,\, (c) \,\, consistency(v_i, v_j) = true \\                
  false, & \text{otherwise}.
\end{cases}
\end{equation} 
Figure \ref{fig:rm_example} illustrates the result of the region merging process after an iterative merging of all RoIs processed. It shows that, based on the predicate in Eq. \ref{eq:reg_merge_pred}, several regions are merged, meanwhile others remain alone as they cannot be merged with other regions (e.g., region number 4).   
The dissimilarity based weight $w_d$ in condition-(b) is related to the statistical properties computed from the regions. In the absence of a boundary among two adjacent regions, one may ignore this condition-(b) and expect similar results, because the condition-(c) could also be used to detect the ineligible regions. However, this will significantly increase the computational time because the eligibility test (Eq. \ref{eq:eligibility}) is significantly faster than applying the RANSAC method.

The \textit{region update} consists of providing an updated representation of the merged region/node.
It is applied immediately after two nodes are identified for merging. We accomplish this by computing the corresponding information ($\pi$, $\mu$, $\kappa$ and $\eta$) of the merged node from the expectation parameters of the individual nodes. For a pair of nodes $v_i$, $v_j$, first we compute the probability ($\pi_{m}$) and expectation parameter ($\eta_{m}$) of the merged node as \cite{hasnat2014wmm, hasnat2015mbhcfmm}:
\begin{equation} 
\label{eq:reg_merge_node}
\pi_{m} = \pi_i + \pi_j \,\, \text{and} \,\,  \eta_{m} = \frac{\pi_i \, \eta_i + \pi_j \, \eta_j}{\pi_{m}}
\end{equation} 
Next, we compute the mean ($\mu_{m}$) and the concentration ($\kappa_{m}$) of the merged node from $\eta_{m}$, see Section \ref{sssec:mvmfmm_dist}. 

The \emph{region merging order} \cite{peng2011automatic} sorts the adjacent regions that should be sequentially evaluated and merged. However, it changes dynamically after each merging occurs. We define the \emph{merging order} using dissimilarity based weights $w_d$ among the adjacent nodes. The adjacent node $v_j$ which has minimum $w_d(v_i, v_j)$ is considered to be evaluated first, e.g., Fig. \ref{fig:rm_example}(c) shows that the region \textit{3} should be evaluated first. We use $w_d$ as the merging order constraint due to its ability to provide a measure of dissimilarity among regions. Such a measure is based on the mean direction ($\mu$) and the concentration ($\kappa$) of the surface normals of the regions. Therefore, with this constraint, the neighboring region, which is most similar w.r.t. $\mu$ and $\kappa$ will be selected as the first candidate to evaluate using Eq. (\ref{eq:reg_merge_pred}).

Algorithm \ref{algo:region_merging} provides the pseudo code for the proposed region merging method. It begins with a set of regions obtained by applying Algorithm \ref{algo:bsc_comb}
on an RGB-D image. As an outcome, it provides the final segmentation result. In the next Section, we evaluate the results obtained from the RGB-D segmentation method detailed in this paper.
\begin{algorithm}
 \SetAlgoLined
 \KwIn{$R = \lbrace r_i \rbrace_{i=1,...,Z} , \,\, G = (V, E), \,\, \kappa_p, \,\, th_b, \,\, th_d \,\, \text{and} \,\, th_r$}
 \KwOut{Final segmentation after region merging.}
 Compute $candidacy(v_i)$ for $\lbrace v_i \rbrace_{i=1,...,Z}$ using Eq. (\ref{eq:candidacy})\;
 Set $i=1$ \;
 \ForEach{$i$}{
 \If{$candidacy(v_i)$ is \emph{true}}{
 \While{no adjacent node of $v_i$ is left to check}{
  Sort the edges $e_{ij}$ (defined with $w_d(v_i, v_j)$ in Eq. \ref{eq:wt_distance_regions}) in an ascending order\;
  Evaluate each $v_{j}$ with the \emph{merging predicate} $P_{ij}$ (Eq. (\ref{eq:reg_merge_pred})) \;
  \uIf{$P_{ij}$ is \emph{true}}{
  	Merge two nodes $v_i$ and $v_j$ and update the RAG\;
  	Start over again from sorting the adjacents $e_{ij}$ of the node $v_i$.
  }
  \Else{ Check the next node }	
  }
  }
 }
 \caption{Region Merging algorithm.}
 \label{algo:region_merging}
\end{algorithm}
\section{Results and Discussion}
\label{sec:results_discussion}
\subsection{Experiments and Results}
\label{ssec:experiments_results}
In this Section, we evaluate the proposed method on the benchmark image database NYUD2 \cite{silberman2012indoor} which consists of 1449 indoor images with RGB, depth and ground-truth information. However, we use the ground truth labels used in \cite{gupta2013perceptual}, because they are corrected for the confused regions, which are labeled as white in the original database. We convert (using MATLAB function) the RGB color information into $L^{*}a^{*}b^{*}$ (CIELAB space) color because of its perceptual accuracy \cite{alata2009there, cheng2011saliency}. For the depth images, we compute the 3D coordinates and surface normals using the toolbox of \cite{ren2012rgb}. 

Our clustering method requires to set initial labels of the pixels and the number of clusters $k$. We initialize it using a combined k-means method with $k=20$. In this k-means method,  the total distance between the cluster center and each observation is computed by adding the Euclidean (for normalized color and 3D positions) and Cosine (for surface normal) distances.
For the region merging we empirically set the thresholds as: $\kappa_{p}=5$ to state that a region is planar, $th_b=0.2$ to state that there is a boundary among two regions, $th_d=3$ to state that there is a dissimilarity between two regions and $th_r=0.9$ to determine the goodness of a plane fitting.

We evaluate performance using standard benchmarks \cite{arbelaez2011contour} which are applied to compare the test and ground truth segmentation: (1) variation of information (\emph{VoI}), it measures the distance between two segmentations in terms of their average conditional entropy; (2) boundary displacement error (\emph{BDE}) \cite{freixenet2002yet}, it measures the average displacement between the boundaries of two segmentations; (3) probability rand index (\emph{PRI}), it measures likelihood of a pair of pixels that has same label; (4) Ground truth region covering (\emph{GTRC}), it measures the region overlap between ground truth and test and (5) Boundary based F-measure (\emph{BFM}), a boundary measure based on precision-recall framework \cite{arbelaez2011contour}. With these criteria a segmentation is better if \emph{VoI} and \emph{BDE} are smaller whereas \emph{PRI}, \emph{GTRC} and \emph{BFM} are larger.

In our experiments, we obtain two sets of segmentation results by using the \textit{Fisher} and \textit{Watson} distribution with \textit{JCSD-RM}. In general, the \textit{Fisher} distribution is the fundamental choice for fitting the normals (subject to unambiguity). In our previous work \cite{hasnat2014unsupervised}, we considered the \textit{Watson} distribution due to the ambiguity in the normals. In this work, yet we consider the \textit{Watson} distribution to study its performance for unambiguous directions. Interestingly, we observe that the results from both \textit{Fisher} and \textit{Watson} distributions are almost equivalent w.r.t. the different measures and computation time. Therefore, in order to avoid redundancy, in this section we do not explicitly present the results of \textit{JCSD-RM} based on the \textit{Watson} distribution. Besides, we compare the results with those obtained from \cite{hasnat2014unsupervised} and observe that the unambiguous normals used in this new paper certainly improves the performance of the overall segmentation task.

We begin the experiments by studying the sensitivity of the proposed method w.r.t. the parameters ($k$, $\kappa_{p}$, $th_b$, $th_d$, $th_r$), which are presented in Table \ref{tab:sensitivity}. The parameter $k$ is related to the clustering method (Section \ref{ssec:jcsa_method}) while $\kappa_{p}$, $th_b$, $th_d$ and $th_r$ are related to the region merging method (Section \ref{ssec:region_merging}). 
From Table \ref{tab:sensitivity}, using the standard deviation of the normalized values of each evaluation metric, we can sort scores in a descending order as: $PRI (0.0057) < VoI (0.0191) < GTRC (0.0198) <  BDE (0.021)$. This means that the BDE measure provides the most discriminating view w.r.t. the parameters and according to it our choice of the threshold values are justified (see Table \ref{tab:sensitivity} where the BDE scores show that the chosen thresholds uniquely provide best results). Moreover, such choice can also be justified using the other measures.
Additional comments about these heuristics based parameters are:
\begin{itemize}
\item
Number of clusters $k$ is inversely related to the number of pixels in a cluster. In segmentation, a smaller $k$ causes a loss of details in the scene, i.e., an under-segmentation, while higher $k$ splits the scene into more regions, i.e., an over-segmentation. 
 Moreover, the computation time of JCSD-RM is proportional to $k$.

\item
We set $\kappa_{p}$ based on the study we did on NYUD2 (see Appendix \ref{apsec:study_planar_stat} for details) which reveals that planar surfaces can be characterized with concentration $\kappa>=5$. While, a lower $\kappa$ value enables to merge non-planar surfaces, a higher value may decrease the probability to merge true planar surfaces.

\item
Following the OWT-UCM \cite{arbelaez2011contour} method, we empirically set the value of $th_b$.

\item
We also set $th_d$ empirically. In theory two regions which have their normals oriented in the same direction should have a negligible Bregman divergence value. However, the inaccurate computation of the shape features and the presence of noise in the acquired depth information often causes the Bregman divergence to be high. From our experience with the images of NYUD2, $th_d$ should be within the range between 2 to 4.

\item
The parameter $th_r=0.9$ is set by following \cite{taylor2013parsing}. Our results in Table \ref{tab:sensitivity} show further justification for this value. 
\end{itemize}
\begin{table*}[h]\footnotesize
\begin{center}
\begin{tabular}{|c|c|c|c|c|c|c|c|c|c|c|c|c|c|c|c|}
\hline
& \multicolumn{3}{|c}{$\{ k, 5, 0.2, 3, 0.9 \}$} & \multicolumn{3}{|c}{$\{ 20, \kappa_{p}, 0.2, 3, 0.9 \}$} & \multicolumn{3}{|c}{$\{ 20, 5, th_b, 3, 0.9 \}$} & \multicolumn{3}{|c|}{$\{ 20, 5, 0.2, th_d, 0.9\}$} & \multicolumn{3}{|c|}{$\{ 20, 5, 0.2, 3, th_r\}$} \\ \hline
                           & 15    & 20      & 25   & 2       & 5          & 8      & 0.1      & 0.2       & 0.3      & 2       & 3          & 4  & 0.85       & 0.9          & 0.95       \\ \hline
\textbf{VoI}       & \textbf{2.17}     & 2.20    & 2.28    & 2.21       & \textbf{2.20}       & 2.27       & 2.34        & \textbf{2.20}      & 2.21        & 2.22       & \textbf{2.20}       & \textbf{2.20}  & \textbf{2.20}       & \textbf{2.20}       & 2.21       \\ \hline
{\textbf{BDE}}  & 9.4     & \textbf{8.97}    & 8.99    & 9.65       & \textbf{8.97}       & 9.08       & 9.25        & \textbf{8.97}      & 9.38        & 9.11       & \textbf{8.97}       & 9.04  & 9.03       & \textbf{8.97}       & 9.03       \\ \hline
\textbf{PRI}                        & 0.90     & \textbf{0.91}    & 0.90    & 0.90       & \textbf{0.91}       & 0.90       & 0.90        & \textbf{0.91}      & 0.90        & \textbf{0.91}       & \textbf{0.91}       & \textbf{0.91} & \textbf{0.91}       & \textbf{0.91}       & \textbf{0.91}       \\ \hline
{\textbf{GTRC}} & \textbf{0.60}     & \textbf{0.60}    & 0.59    & 0.58       & \textbf{0.60}       & 0.58       & 0.56        & \textbf{0.60}      & 0.59        & 0.59       & \textbf{0.60}       & \textbf{0.60} & \textbf{0.60}       & \textbf{0.60}       & \textbf{0.60}      \\ \hline
\end{tabular}
\end{center}
\caption[Sensitivity of JCSD-RM w.r.t. the parameters $\{ k, \kappa_{p}, th_b, th_d, th_r \}$.]{Sensitivity of JCSD-RM with respect to the parameters $\{ k, \kappa_{p}, th_b, th_d, th_r \}$.}
\label{tab:sensitivity}
\end{table*}

Next, we compare the proposed method \textit{\textbf{JCSD-RM}} (joint color-spatial-directional clustering and region merging) with several unsupervised RGB-D segmentation methods such as: RGB-D extension of OWT-UCM \cite{ren2012rgb} (UCM-RGBD), modified Graph Based segmentation \cite{felzenszwalb2004efficient} with color-depth-normal (GBS-CDN), Geometry and Color Fusion method \cite{dal2012fusion} (GCF) and the Scene Parsing Method \cite{taylor2013parsing} (SP). For the UCM-RGBD method we obtain best score with threshold value 0.1. The best results from GBS-CDN method are obtained by using $\sigma=0.4$. To obtain the optimal multiplier ($\lambda$) in GCF \cite{dal2012fusion} we exploit the range 0.5 to 2.5. For the SP method, we scaled the depth values (1/0.1 to 1/10 in meters) to use author's source code \cite{taylor2013parsing}.

Table \ref{tab:accuracy_eval} presents (best appears as bold) the comparison w.r.t. the average score of the benchmarks. Results show that JCSD-RM performs best according to PRI, VoI,  GTRC  and BDE. Moreover, it is comparable according to BFM. The reason is that, BFM favors methods like UCM-RGBD which is specialized in contours detection. 
On the other hand, JCSD clustering method provides an approximation (see e.g., Figure \ref{fig:rm_predicates}) of the object boundary which is often coarse. This can be improved by developing a spatially constrained clustering method, such as \cite{nguyen2013fast}. A better boundary approximation will subsequently improve the performance of the RM method.
Therefore, we can say that JCSD-RM could be further improved by incorporating the boundary information more efficiently.
\begin{table}[h]
\begin{center}
\begin{tabular}{|l|l|l|l|l|l|}
\hline
\multicolumn{1}{|c|}{} & \textbf{VoI}           & \textbf{BDE}           & \textbf{PRI}   & \textbf{GTRC} & \textbf{BFM}
\\ \hline
\textbf{UCM-RGBD}                         & 2.35          & 9.11 & 0.90         & 0.57 & \textbf{0.63}         \\ \hline
\textbf{GBS-CDN}                             & 2.32          & 13.23         & 0.81          & 0.49         & 0.53          
\\ \hline
\textbf{GCF}                             & 3.09          & 14.23         & 0.84          & 0.35         & 0.42         
\\ \hline
\textbf{SP}                            & 3.15          & 10.74          & 0.85          & 0.44         & 0.50         
\\ \hline
\textbf{JCSD}                      & 2.68 & 10.00        & 0.87  & 0.46         & 0.46 \\ \hline
\textbf{JCSD-RM}                      & \textbf{2.2} & \textbf{8.97}         & \textbf{0.91}  & \textbf{0.6}         & 0.61 \\ \hline
\end{tabular}
\end{center}
\caption[Comparison with the state of the art.]{Comparison with the state of the art. Methods: \textbf{UCM-RGBD} \cite{ren2012rgb}, \textbf{GBS-CDN} \cite{felzenszwalb2004efficient}, \textbf{GCF} \cite{dal2012fusion}, \textbf{SP} \cite{taylor2013parsing}, \textbf{JCSD} and \textbf{JCSD-RM} (proposed). \textbf{Boldface} indicates the best results.}
\label{tab:accuracy_eval}
\end{table}

\begin{figure}[H]
\centering
\includegraphics[scale=0.53]{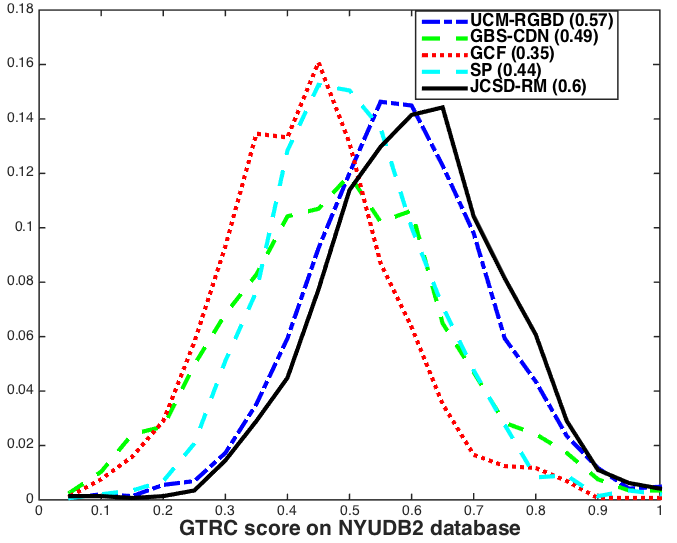}
\caption[Histogram of GTRC scores of different methods.]{Histogram of \emph{GTRC} \cite{arbelaez2011contour} scores of different methods.}
\label{fig:gtrc_analysis_nyudb}
\end{figure}

\emph{Ground Truth Region Covering (GTRC)} has been chosen as one of most prominent measure of evaluation for segmentation methods \cite{arbelaez2011contour, gupta2013perceptual}. In Table \ref{tab:sensitivity} and \ref{tab:accuracy_eval}, we observed that it provides discriminative score to evaluate and differentiate among the different state-of-the-art methods. Fig. \ref{fig:gtrc_analysis_nyudb} provides further analysis on NYUD2 \cite{silberman2012indoor} using histograms of GTRC scores. We observe that, while the JCSD-RM and UCM-RGBD covers quite similar regions in the histogram, others are significantly different especially in the higher GTRC region. Particularly, the JCSD-RM has lower percentage of images with low GTRC score region and higher percentage for the high GTRC score region. 
In order to conduct the experiments we used a 64 bit machine with Intel Xenon CPU and 16 GB RAM. The JCSD-RM method is implemented in MATLAB, which on average takes 38 seconds, where 31 seconds for the clustering and 7 seconds for region merging. In contrast, UCM-RGBD (MATLAB and C++) takes 110 seconds. Therefore, JCSD-RM is $\approx$3 times faster\footnote{To perform a fair comparison, we conducted this experiment with half scaled image. This is due to the fact that the computational resource did not support to run UCM-RGBD for the full scale image.} than UCM-RGBD. Moreover, we believe that implementing JCSD-RM in C++ will significantly reduce the computation time.

To further analyze the computation time of JCSD-RM, we run it for different image scales. Table \ref{tab:jcsa_rm_comp_scale} presents relevant information from which we see that the reduction rate of JCSD computation time (in sec) w.r.t. different scales is approximately equivalent to the reduction rate of the number of pixels.
\begin{table}[h]\footnotesize
\begin{center}
\begin{tabular}{|l|l|l|l|l|}
\hline
\textbf{Scale}       & \textbf{1} & \textbf{1/2} & \textbf{1/4} & \textbf{1/8} \\ \hline
\textbf{Num. pixels} & 239k       & 60k          & 15k          & 4k           \\ \hline
\textbf{JCSD} (req. time in sec)        & 132        & 31           & 8            & 1.5          \\ \hline
\textbf{RM} (req. time in sec)          & 42         & 7            & 1.4          & 0.33         \\ \hline
\end{tabular}
\end{center}
\caption{Computation time of JCSD-RM w.r.t. different image scales.}
\label{tab:jcsa_rm_comp_scale}
\end{table}
\subsection{Discussion}
\label{ssec:discussion}
\begin{figure*}[h]
\centering
\centerline{\includegraphics[scale=0.27]{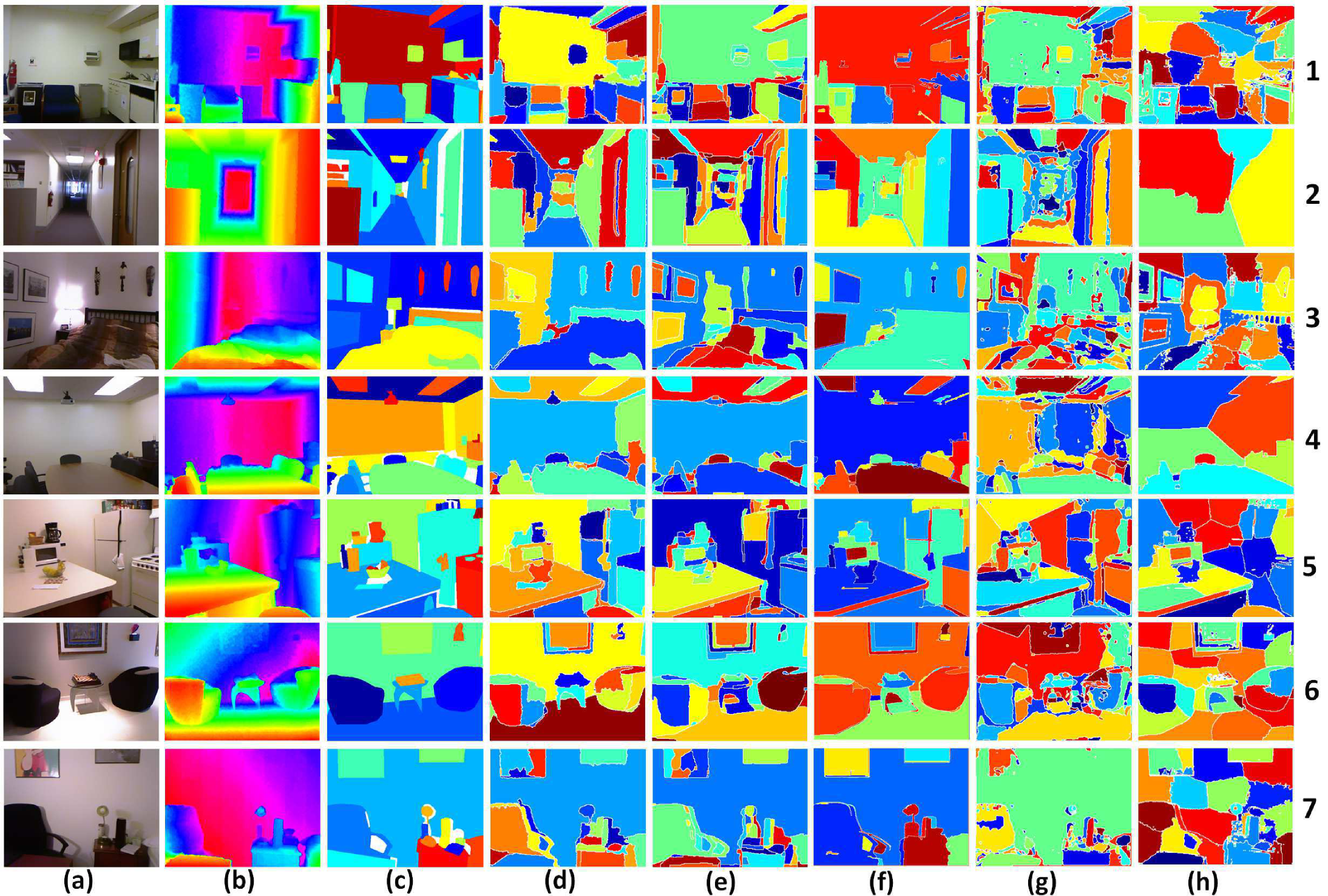}}
\caption[Segmentation examples on NYU RGB-D database using different methods.]{Segmentation examples (from top to bottom) on NYU RGB-D database (NYUD2). (a) Input Color image (b) Input Depth image (c) Ground truth (d) JCSD-RM (\emph{our proposed}) (e) UCM-RGBD \cite{ren2012rgb} (f) GBS-CDN \cite{felzenszwalb2004efficient} (g) SP \cite{taylor2013parsing} and (h) GCF \cite{dal2012fusion}.}
\label{fig:rag_example}
\end{figure*}

Several segmentation results are illustrated in Fig \ref{fig:rag_example}. These examples confirm that the segmentation from JCSD-RM (our proposed) and UCM-RGBD are competitive. However, they let us note several differences: (a) JCSD-RM is better in providing the details of indoor scene structures whereas UCM-RGBD loses them sometimes (see ex. rows 3 to 5); (b) UCM-RGBD provides better estimation of the object boundaries whereas JCSD-RM gives a rough boundary and (c) UCM-RGBD shows more sensitivity on color whereas JCSD-RM is more sensitive on directions. The GBS-CDN method provides visually pleasing results, however it often tends to loose details (see ex. rows 1 to 4) of the scene structure (e.g., merges wall with ceiling). Results from the SP method seems to be severely sensitive to the varying illumination and rough changes in surfaces (see ex. row 3). The GCF method performs over-segmentation (see ex. rows 1, 3, and rows 5-7) or under-segmentation (see ex. rows 2 and 4), which is a drawback of such algorithm as it is often unable to estimate the correct number of clusters in real data. Moreover, the GCF method often fails to discriminate major surface orientations (see ex. rows 1, 2 and 4) as it does not consider the direction of surfaces (normals).

\begin{figure*}[h]
\centering
\includegraphics[viewport = 80 150 1190 560, clip = true, scale=0.42]{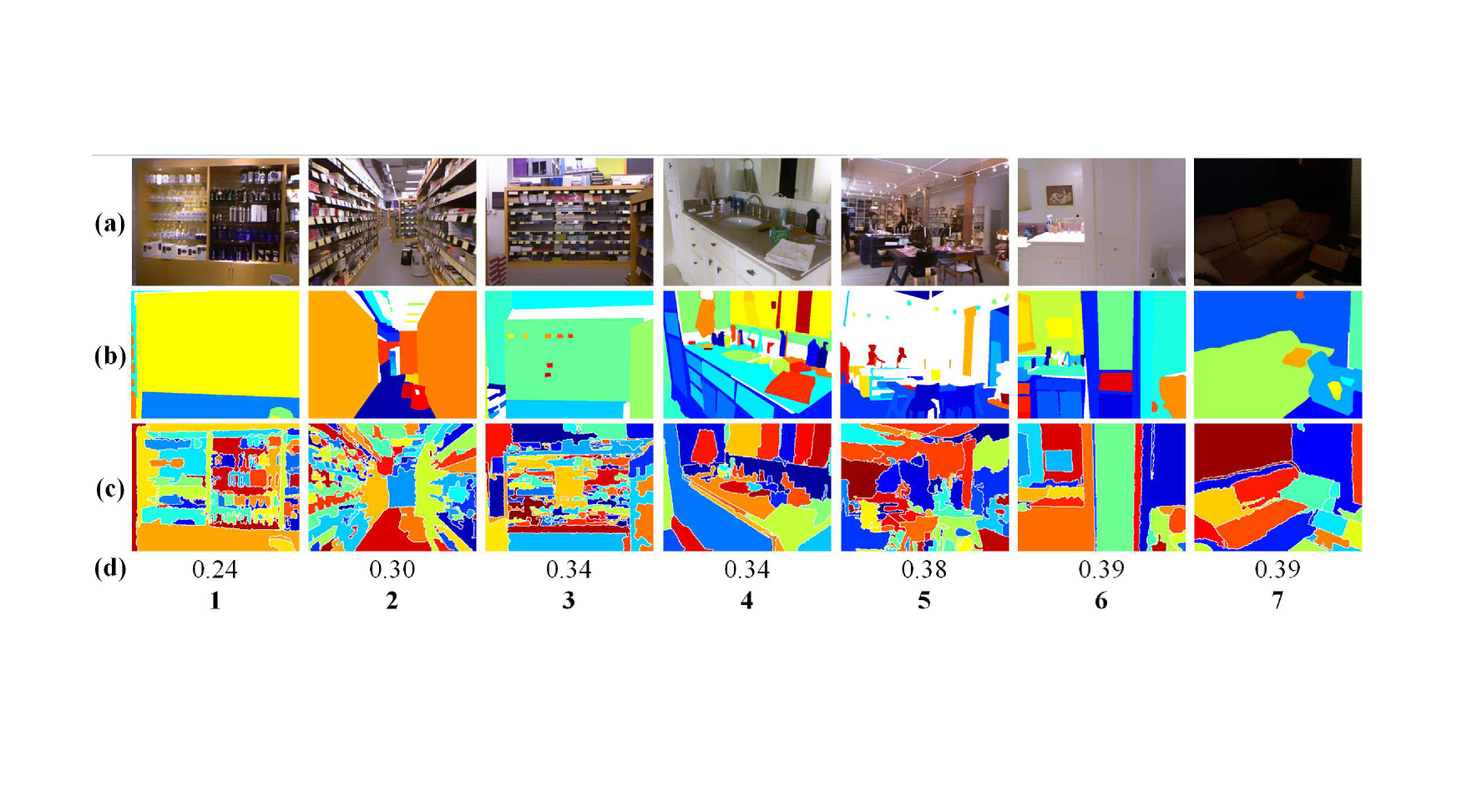}
\caption[Segmentation examples with lower GTRC scores.]{Segmentation examples with lower GTRC scores (less than 0.4). (a) Input Color Image (b) Ground Truth Segmentation (c) Segmentation with the JCSD-RM method and (d) GTRC score.}
\label{fig:gtrc_visual_nyudb}
\end{figure*}

Now, in Fig. \ref{fig:gtrc_visual_nyudb} let us focus and analyze some segmentation examples which have lower (less than 0.4) GTRC score. Average GTRC score of JCSD-RM is 0.6 (see Table \ref{tab:sensitivity} and \ref{tab:accuracy_eval}). Results show several cases for low scores:
\begin{itemize}
\item
JCSD-RM method tends to provide more details (over-segment) while the ground truth keeps minimum details, see ex. columns 1 to 3, and 5 in Fig. \ref{fig:gtrc_visual_nyudb}. 
\item
JCSD-RM method does not provide enough details (under-segment) while the ground truth does, see ex. columns 4 and 6 in Fig. \ref{fig:gtrc_visual_nyudb}. This is a very difficult case, as looking at the images we can see that the under-segmented regions have similar color, depth and normal which in a general case is difficult to segment without additional knowledge.
\item
Example column 7 shows a characteristic example of JCSD-RM, which is to be biased on surface normals. This causes the furniture (sofa) to be segmented into several parts. Perhaps this can be improved by incorporating color based merging heuristics in our region merging method.
\end{itemize}

Now, let us focus particularly on the JCSD method, which is based on a statistical image generation model defined from the na\"{\i}ve Bayes assumption \cite{zhang2004optimality, lewis1998naive}. Computer vision or data analysis experts may question about the assumption of independence between the color, depth and normal. While we partially agree with the experts, in a first step we made this assumption because of two reasons: (a) propose a simplified method to understand the underlying grouping mechanisms of different image features in a combined fashion and (b) to empirically verify (in a second step) the relevance of this assumption in an unsupervised context. 

\begin{table}[h]
\begin{center}
\begin{tabular}{|c|c|c|c|c|}
\hline
                     & \textbf{VoI} & \textbf{BDE} & \textbf{PRI} & \textbf{GTRC} \\ \hline
\textbf{JCSD}        & \textbf{2.68}         & \textbf{10.00}        & \textbf{0.87}         & \textbf{0.46}          \\ \hline
\textbf{All\_GMM}         & 3.01         & 11.04        & \textbf{0.87}         & 0.43          \\ \hline
\textbf{PCA\_GMM}     & 3.94         & 12.01        & 0.85         & 0.34          \\ \hline
\textbf{Ind\_GMM}    & 3.22         & 11.04        & 0.86         & 0.41          \\ \hline
\textbf{JCSD+RM}     & \textbf{2.21}         & \textbf{8.97}         & \textbf{0.91}         & \textbf{0.60}          \\ \hline
\textbf{All\_GMM + RM}    & 2.41         & 9.28         & 0.90         & 0.58          \\ \hline
\textbf{PCA\_GMM+RM}  & 2.85         & 10.25       & 0.88          & 0.50          \\ \hline
\textbf{Ind\_GMM+RM} & 2.62         & 10.15        & 0.89         & 0.53          \\ \hline
\end{tabular}
\end{center}
\caption{Comparison among different image models and the clustering results with/without the region merging method. \textbf{Boldface} indicates the best results among similar methods (with and without RM).}
\label{tab:comp_image_models}
\end{table}

Table \ref{tab:comp_image_models} provides results w.r.t. several alternative models. First, we consider a unified model, called All\_GMM, that fits a Gaussian Mixture Model (GMM) for all features together, i.e., no independence among features. Results show that, while JCSD is same with All\_GMM only for PRI measure, it is better w.r.t. the VoI, BDE and GTRC measures. In terms of average computing time, JCSD takes 31 sec and All\_GMM takes 45 sec, i.e., JCSD is 1.45 times faster. Moreover, in numerous images, All\_GMM fails due to the ill-conditioned covariance matrix. Based on this observation, we consider a different model, called PCA\_GMM, which reduces the features using PCA method by considering 95\% variances of the data. On average the reduced feature dimension was 7. Results show that, while the performance decreases remarkably, there was no potential gain on computational time. From these results, we can see that our simplified image model is better w.r.t. the standard measures and computation time.

Another JCSD clustering related issue is the assumption of a known maximum number of clusters $k_{max}$. In the context of mixture model based clustering, numerous methods exist to automatically select the number of clusters. For example, \cite{fraley2007model} used the Bayesian Information Criteria (BIC), \cite{figueiredo2002unsupervised} employed the Minimum Message Length (MML), \cite{biernacki2000assessing} proposed the Integrated Completed Likelihood (ICL) and \cite{alata2009there} applied $\Phi_{\beta}$ criterion. Besides these criteria, in our previous work \cite{hasnat2015mbhcfmm}, we proposed a modification of the slope heuristic based method. For this work, we developed and experimented (not presented in this paper) all of the above-mentioned methods and observed a large number of over-segmented and under-segmented images. We realized that, while it is difficult to improve an under-segmentation, it is easier to improve an over-segmentation, e.g., by using a region merging method. Therefore, we decided to avoid the idea of automatic number of clusters selection and use the notion of a predefined $k_{max}$.

Now, let us focus on a different concern related to the use of directional (Fisher or Watson) distribution for the surface normals. An expert could easily argue that the Gaussian distribution can be used in the place of directional distribution. Although, in our previous work \cite{hasnat2015mbhcfmm} on clustering normals, we have shown that directional distribution is more appropriate than Gaussian distribution, here we provide further verification within the context of joint clustering with independent assumption. We consider JCSD type model, called Ind\_GMM, which replaces the directional distribution with the Gaussian distribution. Results show that, JCSD is better w.r.t. all evaluation measures, which further demonstrates the efficiency and relevance of our proposed image model. Interestingly, we observe that Ind\_GMM provides slightly lower performance than All\_GMM, which reveals that the independence assumption is context dependent and does not necessarily provide better results if the unsupervised classifier (here GMM) remains same and may provide better results if we understand the heterogeneous (i.e., combination of different types of features) data and build a model to handle them appropriately.

Comparing JCSD with JCSD-RM (Table \ref{tab:accuracy_eval} and \ref{tab:comp_image_models}), we can decompose the contributions of \emph{clustering} and \emph{region merging} in JCSD-RM. We see that \emph{region merging} improves clustering output from 0.46 to 0.6 (30.43$\%$) in GTRC. 
We believe that JCSD-RM can be improved and extended further in the following ways:
\begin{itemize}
\item
Including a pre-processing stage, which is necessary because the shape features are often computed inaccurately due to noise and quantization \cite{barron2013intrinsic}. Moreover, we observed significant noise in NYUD2 color images which were captured especially in low light condition. A method like Scene-SIRFS (shape, illumination and reflectance from shading) \cite{barron2013intrinsic}, which recover the intrinsic scene properties, can be used for pre-processing purpose.
\item
Enhancing the clustering method by adding contour information \cite{arbelaez2011contour} efficiently. Additionally, we may consider spatially constrained model such as \cite{nguyen2013fast} which incorporates boundary information by adding spatially varying constraints in the clustering task.
\item
Enhancing the region merging method with color information. To this aim, we can exploit the estimated reflectance information (using \cite{barron2013intrinsic}), such that the varying illumination is discounted. 
\end{itemize}
\section{Conclusion}
\label{sec:conclusion}
We proposed an unsupervised indoor RGB-D scene segmentation method. Our method is based on a statistical image generation model, which provides a theoretical basis for fusing different cues (e.g., color and depth) of an image. In order to cluster w.r.t. the image model, we developed an efficient joint color-spatial-directional clustering method based on Bregman divergence. Additionally, we proposed a region merging method that exploits the planar statistics of the image regions. We evaluated the proposed method with a database of benchmark RGB-D images and using widely accepted evaluation metrics. Results show that our method is competitive w.r.t. the state of the art and opens interesting perspectives for fusing color and geometry. We foresee several possible extensions of our method: more complex image model and clustering with additional features, region merging with additional hypothesis based on color. Moreover, we believe that the methodology proposed in this paper is equally applicable and extendable for other complex tasks, such as joint image-speech data analysis.

\appendices
\section{Bregman Divergence (BD) - an alternative formulation and relationship}
\label{ssec:bd_efd}
A multivariate probability density function $f(\textbf{x}|\theta)$ belongs to the regular exponential family if it has the following form \cite{liu2012shape, banerjee2005clusteringb}:
\begin{equation} 
\label{eq:efd_2}
f\left(\textbf{x}|\theta\right ) = \text{exp}\left (\left \langle t(\textbf{x}), \theta ) \right \rangle -F(\theta) + k(\textbf{x})\right )
\end{equation} 
Here, 
$t(\textbf{x})$ is the sufficient statistics, 
$\theta$ is the natural parameters,
$F(\theta)$ is the log normalizing function,
$k(\textbf{x})$ is the carrier measure and
$<.,.>$ is the inner product.  

The expectation of the sufficient statistics $t(\textbf{x})$ is called the expectation parameter, $\eta=E[t(\textbf{x})]$. There exists a one-to-one correspondence between expectation $(\eta)$ and natural $(\theta)$ parameters, which exhibits dual relationships among the parameters and functions as \cite{banerjee2005clusteringb}:
\begin{equation}
\label{eq:eta_theta_rel}
\eta = \nabla F(\theta) \;\;\; and \;\;\; \theta = (\nabla F)^{-1} (\eta)
\end{equation} 
and
\begin{equation}
\label{eq:grad_dual_ln}
G(\eta) = \left \langle (\nabla F)^{-1} (\eta), \eta \right \rangle - F\left ( (\nabla F)^{-1} (\eta) \right )
\end{equation}
Here, $\nabla F$ is the gradient of  $F$. $G(.)$ is the Legendre dual of the log normalizing function $F(.)$. See details in Section 3.2 of \cite{banerjee2005clusteringb}.

For a strictly convex function $F(.)$, Bregman Divergence, $D_{F}(\theta_{1}, \theta_{2})$ can be formally defined as \cite{banerjee2005clusteringb}:
\begin{equation}
\label{eq:BD_theta}
D_{F}\left ( \theta_{1}, \theta_{2} \right )= F(\theta_{1}) - F(\theta_{2}) - \left \langle \theta_{1} - \theta_{2}, \nabla F(\theta_{2}) \right \rangle
\end{equation}
$D_F(\theta_1, \theta_2)$ measures the distance using the tangent function at $\theta_2$ to approximate $F$. This can be seen as the distance between the first order Taylor approximation to $F$ at $\theta_2$ and the function evaluated at $\theta_1$ \cite{liu2012shape}. 
The one-to-one correspondence in Eq. (\ref{eq:eta_theta_rel}) provides the dual form of BD (of Eq. (\ref{eq:BD_theta})) as:
\begin{equation}
\label{eq:BD_eta}
D_{G}\left ( \eta_{1}, \eta_{2} \right )= G(\eta_{1}) - G(\eta_{2}) - \left \langle \eta_{1} - \eta_{2}, \nabla G(\eta_{2}) \right \rangle
\end{equation}

Due to the bijection between BD and the Exponential families, Eq. (\ref{eq:BD_theta}) and (\ref{eq:BD_eta}) can be used to measure the dissimilarity between distributions of the same family. The bijection is expressed as: $f(\textbf{x}|\theta)=\text{exp}(-D_{G}(t(\textbf{x}),\eta))J_{G}(\textbf{x})$ where $J_{G}$ is a uniquely determined function. We used this formulation in Eq. (\ref{eq:efd}) of this paper. For more details, see Theorem 3 of \cite{banerjee2005clusteringb}.

Bregman divergences (BD) generalize the squared Eucilidean distance, Mahalanobis distance, Kullback-Leibler divergence, Itakura-Saito divergence etc. See Table 1 of \cite{banerjee2005clusteringb} and \cite{boissonnat2010bregman} for a list and corresponding $D_F(.,.)$. Besides, BD has the following interesting properties \cite{boissonnat2010bregman}:

\begin{itemize}
\item Non-negativity: The strict convexity of $F$ implies that, for any $\theta_1$ and $\theta_2$, $D_F(\theta_1, \theta_2) \geq 0$ and $D_F(\theta_1, \theta_2) = 0$ if and only if $\theta_1=\theta_2$.
\item Convexity: Function $D_F(\theta_1, \theta_2)$ is convex in its first argument $\theta_1$ but not necessarily in the second argument $\theta_2$.
\item Linearity: BD is a linear operator, i.e., for any two strictly convex functions $F_1$ and $F_2$ and $\lambda \geq 0$:
\begin{equation*}
D_{F1+\lambda F2}(\theta_1, \theta_2) = D_{F1}(\theta_1, \theta_2) + \lambda \; D_{F2}(\theta_1, \theta_2)
\end{equation*}
\end{itemize}
\section{Study of planar statistics}
\label{apsec:study_planar_stat}
\begin{figure}[H]
\centering
\includegraphics[scale=0.55]{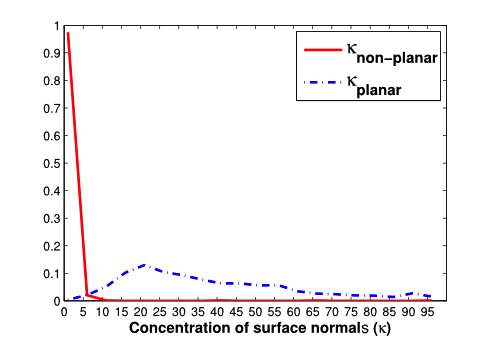}
\caption{Histogram of $\kappa$ values for planar and non-planar surfaces.}
\label{fig:hist_pl_nplan}
\end{figure}

For this study we applied clustering (with Fisher Mixture Model \cite{hasnat2015mbhcfmm}) on surface normals of each image of the NYU Depth database V2 (NYUD2) \cite{silberman2012indoor}. Fig. \ref{fig:hist_pl_nplan} illustrates the histograms of $\kappa$ (concentration of surface normals) values for planar and non-planar surfaces. These histograms have been obtained from an analysis of four category of segmented surfaces: (1) planar; (2) non-planar ; (3) planar + non-planar and (d) unknown (category not sure). 
We use the NYUD2 dataset as it provides labeled 3D images of indoor scenes. A total of 5410 RoIs were analyzed, among them 2559 represented planar surface meanwhile 793 represented non-planar surfaces.
Then we computed the histogram of $\kappa$ values for all these RoIs.
We observed that 99.88$\%$ of planar surfaces has $\kappa>5$ and 99.5$\%$ of non-planar surfaces has $\kappa<5$. 
This heuristically shows that the assumption about planar statistics used in Eq. (\ref{eq:candidacy}), based on $\kappa$ values, is appropriate for region merging.

\bibliographystyle{IEEEtran}
\bibliography{pami_bib}
\end{document}